%% file: ms.tex
\documentclass[10pt,twocolumn,letterpaper]{article}

\usepackage{titling}
\usepackage{iccv}
\usepackage{times}
\usepackage{epsfig}
\usepackage{graphicx}
\usepackage{amsmath}
\usepackage{amssymb}
\usepackage{scrextend}
\usepackage{multirow}
\usepackage[english]{babel}
\usepackage{subcaption}
\usepackage{paralist, tabularx}
\usepackage{amsthm}
\usepackage{enumitem}

\newtheorem{theorem}{Theorem}

\newcommand{\vS}{{\mathbb{S}^2}}
\newcommand{\vM}{{\mathbb{M}}}
\newcommand{\vx}{{\mathbf{x}}}

\newcommand{\R}{\mathbb{R}}
\newcommand{\vn}{\mathbf{n}}
\newcommand{\vh}{\mathbf{h}}
\newcommand{\vi}{\mathbf{i}}
\newcommand{\vo}{\mathbf{o}}

\usepackage[pagebackref=true,breaklinks=true,letterpaper=true,colorlinks,bookmarks=false]{hyperref}

\iccvfinalcopy 

\def\iccvPaperID{****} 

\usepackage{mdframed}

\begin{document}

\title{One Ring to Rule Them All: a simple solution to multi-view 3D-Reconstruction of shapes with unknown BRDF via a small Recurrent ResNet}

\author{Ziang Cheng\textsuperscript{1}, Hongdong Li\textsuperscript{\rm 1}, Richard Hartley\textsuperscript{\rm 1}, Yinqiang Zheng\textsuperscript{\rm 2}, Imari Sato\textsuperscript{\rm 3}\\
\textsuperscript{\rm 1}Australian National University\\\textsuperscript{\rm 2}The University of Tokyo,\textsuperscript{\rm 3}National Institute of Informatics, Japan\\
{\tt\small \{ziang.cheng,hongdong.li\}@anu.edu.au}}

\maketitle

\begin{abstract}
\input{ICCV21_abstract} 
\end{abstract}

\section{Introduction}
Reconstructing the 3D shape of object or scene from their multi-view images is one of the central tasks in computer vision research. Traditional multi-view 3D-reconstruction methods often assume that the objects or scenes of interest are largely diffuse (\ie close to Lambertian) or texture-rich, therefore, allowing for reliable cross-view image correspondences. However, in reality, many commonly seen objects are made of generic materials possibly with glossy, metal-like appearances, violating the brightness-constancy assumption needed for establishing image correspondences. 

It remains an open challenge to estimate the 3D geometry of objects of unknown arbitrary materials. Furthermore, when the object is illuminated by a moving (active) light source, it further complicates the task, because the visual appearance of a non-Lambertian object is not only view-dependent but also light-dependent in general.  Very few approaches have been attempted at this challenging  task. The only existing ones, based on photometric stereo, are plagued by solving difficult mathematical optimization problems often involving a highly non-convex objective function derived from photometric image rendering equation. As a result, their performances critically depends on the quality of the initialization or require significant manual intervention (\eg parameter tuning \cite{park2016robust,nam2018practical}).

In this paper, we propose a neural-network based feature-correspondence-free method for reconstructing both the shape of an object and its spatially-varying reflectance model in the form of a BRDF (Bidirectional Reflectance Distribution Function).  This allows novel synthetic views of the object to be rendered with high realism.

\begin{figure*}[t!p]  \centering
\includegraphics[width=0.6\linewidth]{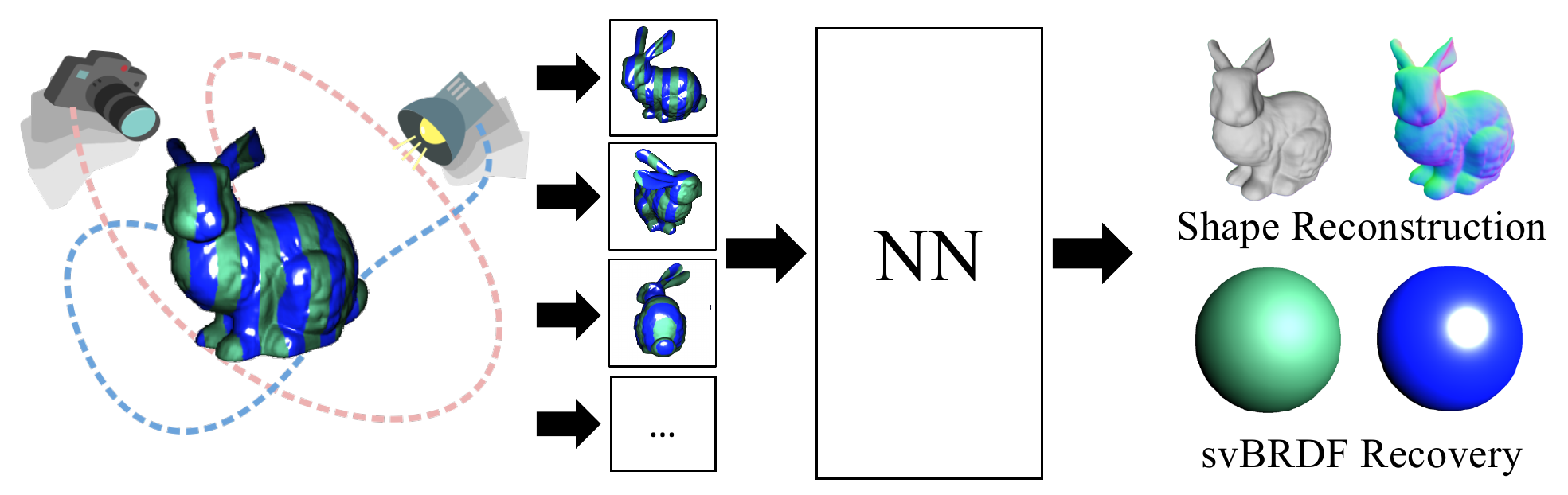}
\caption{\small Problem Setting: an object with unknown surface materials (which can be spatially-varying non-Lambertian,  parameterized by {\small svBRDF} function) illuminated by a possibly moving light source, and observed by a moving camera.  We develop a simple neural-network based solution that effectively recovers high-quality shape and the {\small svBRDF} accurately and reliably from its multi-view image observations.}
\label{fig:setup}
\end{figure*}

Key to our method is the representation of object shape by a smooth vector field on the ambient space $\R^3$ along which a canonical shape ``flows'' to the desired shape.  We show mathematically that the resulting shape must 
be a smooth embedding of a sphere, and that all genus-zero shapes can be represented in this way. The vector field (\ie a smooth mapping from $\R^3$ to $\R^3$) is computed by a single MLP (Multi-layer perceptron), integrated via a recurrent architecture with a ``ring" (feedback) connection.  This cascade is implemented by a recursive residual network, which we call the {\em Shape-Net}. The input to our method is a set of views of an object for which the positions of the camera and the light source are known.  The output consists of a watertight shape and {\small BRDF} parameters for each point on the surface, both embedded in the same sphere. Such a surface representation differs from explicit mesh or implicit level set used in many previous works  \cite{wang2018pixel2mesh,jiang2020sdfdiff}.  Given the success of this approach, we believe that this method of defining a shape via a vector field will have wide applications beyond $3D$-reconstruction. Representation of diffeomorphisms using flows has been used before in computer vision \cite{LMLDD}, but not with deep neural nets.  

By extensive experiments, we demonstrate that our method produces accurate and compelling shape and svBRDF reconstructions, even without initialization. In the sequel, we will first describe the problem setting, our new method and its theory, followed by experiment validations.  We defer ``Related work" to a later section for the sake of smooth reading. 

\section{Problem Setting and Formulation} 
\subsection{Problem Setting}
Consider a very general multi-view imaging setup, where the 3D object to be reconstructed may have unknown, generic (\eg non-Lambertian), and possibly spatially-varying (even per-point different) surface reflectance.  The object surface can be entirely smooth or texture-less, or be coated with different paints or materials. The light source can be either near-field or distant, and both the light and the camera are allowed to move freely between image shootings. We assume that the poses of the camera and light-source position are pre-calibrated, though we do not constrain their relative positions. 

Fig-\ref{fig:setup} (left) depicts the concept of our problem setting, where a smooth and partially shiny object is photographed by a moving camera under a moving light source.  The task is to recover the 3D shape of the object, as well as per-point surface reflectance (parametrized by a spatially-varying BRDF, or svBRDF).  Traditional multi-view SFM/MVS methods (such as ColMAP \cite{schoenberger2016sfm}, PMVS \cite{furukawa2009accurate} \etc) are unsuitable to handle such a general imaging setup due to the violation of the color constancy or Lambertian assumption.

\subsection{Shape parametrization} 
To ease exposition, we shall hereafter assume that the object has a bounded surface of {\em genus-0 topology}. In other words, it is a closed 2-manifold surface having no hole, hence is topologically equivalent to a unit 2-sphere.~~ For all genus-0 surfaces, the unit 2-sphere ($\vS$) is the most natural parametrization domain since there always exists a smooth and invertible mapping (\ie diffeomorphism) between the unit sphere and any smooth 2-manifold of genus-0. Such a mapping is called the {\em spherical embedding}.

In the context of multi-view 3D shape reconstruction, this spherical embedding offers a convenient way to to encode object surface {\em a-priori} -- namely, even before the object surface is reconstructed. 

To better see this, let us use $\vM$ to denote the object surface manifold, and use $\vx$ to denote a 3D point on the manifold. Suppose a diffeomorphism $\Phi:\vS\mapsto\vM$ is established between the sphere and the manifold, we have $\Phi(\mathbf{s})=\vx$, where $\mathbf{s}$ denotes a point on the unit sphere.  Finding a spherical embedding is equivalent to saying that we have reconstructed the shape. This is because, feeding all points on the unit sphere to $\Phi$ will trace out the entire 3D surface, \ie$~\Phi(\mathbb{S}^2)\rightarrow\mathbb{M}$. The diffeomorphism property also ensures $\Phi$ is differentiable, therefore a surface normal always exists on the target manifold.

\begin{mdframed}[linewidth=1pt]
We convert the task of shape reconstruction to {\bf learning a diffeomorphism} $\Phi$, defined by the flow on
a vector field, conditioned on the multi-view input images. We develop a simple neural-net (\ie Shape-Net) to learn this map.
\end{mdframed}

At first glance, the above genus-0 assumption may seem restrictive. However, we note that (1) in practice our method can be easily extended to objects of higher genus, by embedding their surface in a suitable {\em canonical} domain (\eg visual hull) (2) our method can still approximate higher genus shapes even from a genus-0 embedding (The reader is referred to the Appendix for an ablation test). 
\subsection{BRDF Parametrization}

The {\small{BRDF}} function describes surface reflectance as a 4D function of the incident light and outgoing viewing directions relative to surface normal at a surface point. There are various ways to parametrize a {\small{BRDF}}. For simplicity, in this paper, we use a physically-based Cook-Torrance model \cite{cook1982reflectance} $B(\vn,\vi,\vo;\theta)$, defined for surface normal $\vn$, and incident and viewing directions $\vi,\vo$ as
\begin{equation}
{\small  B(\vn,\vi,\vo;\theta) = \rho^{rgb} + \rho^\gamma \frac{D(\vn,\vh;r)FG(\vn,\vi,\vo)}{\pi (\vn\cdot\vi)(\vn\cdot\vo)},} 
\end{equation}
where $\rho^{rgb}$ and $\rho^\gamma$ are the diffuse RGB and specular albedo, and $r\in(0,1)$ defines the roughness of the material. $\vh$ is the half-vector between incident and viewing rays $\vi,\vo$. $D$ defines the angular distribution of specular highlights, and $G$ is a mask-shadowing term. Under our imaging condition the Fresnel term $F$ is a constant hence can be omitted (\cf\cite{nam2018practical}). This way, we are able to encode a {\small{BRDF}} by a compact 5D vector of $\theta=\{\rho^r,\rho^g, \rho^b, \rho^\gamma, r\}$ \cite{cook1982reflectance}.

It is worth mentioning that, our method is not married to any particular choice of {\small{BRDF}} models. Here we use the Cook-Torrance model only for its compactness. In fact, it can be trivially adapted to other forms of BRDF models (\eg \cite{walter2007microfacet,nishino2009directional,Matusik2003A,nielsen2015optimal,chen2020invertible}) as long as the model is differentiable. Let us use $\Psi:\vS\mapsto\mathbb{R}^5$ to denote a function that maps any point on the unit sphere to the 5D Cook-Torrance vector, then we have 
\begin{equation}
    \Psi\circ\Phi^{-1}(\vx)=\theta(\vx)\in \mathbb{R}^5,
\end{equation}
 where $\theta(\vx)$ is the 5D {\small{BRDF}} parameters at surface point $\vx$, and $\Phi$ is the diffeomorphism between the unit sphere and the object surface. Note that once the spherical embedding $\Phi$ is given, we do not need to know its inverse $\Phi^{-1}$ as long as it exists, since we can index $\vx$ by $\mathbf{s}$ instead.
 
\begin{mdframed}[linewidth=1pt]
We reduce the task of {\small{BRDF}} estimation to {\bf learning a map $\Psi:\vS\mapsto\mathbb{R}^5$} from the unit sphere to a 5D space, conditioning on the multi-view input images. We employ a plain 6-layer MLP (BRDF-Net) for this task.
\end{mdframed}
Because both the $\Phi$ and $\Psi$ share the same spherical domain, and are both conditioned on the same set of multi-view input images, they need to be solved jointly. Traditional optimization often adopt an alternated procedure to solve such a {\em chicken-and-egg} problem, usually from a sufficiently good initialization. In this paper, we show however, one can easily solve this hard problem by training two simple neural nets, from {\bf random initialization}.

\section{Method}
\begin{figure}[t!h!] 
    \includegraphics[width=0.5\textwidth]{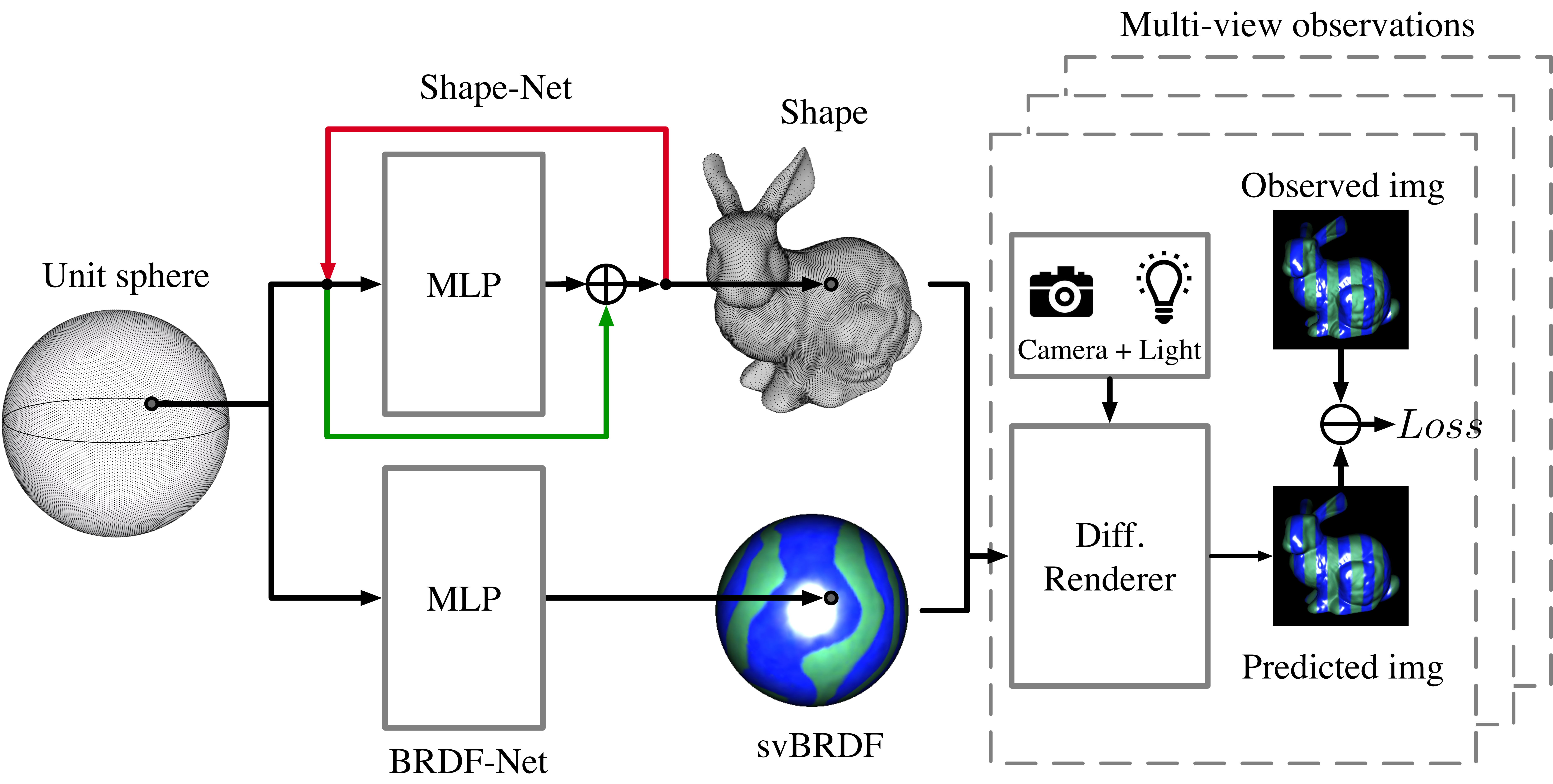} 
    \caption{\small Overall pipeline of our method. The left half of the graph depicts the Shape-Net and {\small{BRDF}}-net. When input points have traversed the entire unit sphere, the outputs of the Shape-Net will trace out a complete shape. The outputs of the {\small{BRDF}}-Net yield a full set of {\small{svBRDF}} estimates. Feeding the above predicted shape and {\small{svBRDF}} into the differentiable renderer, along with the corresponding camera pose and light position, generates a predicted image. Comparing this image with the actual image produces the training loss to train the networks.\label{fig:neural_networks_concept}}
\end{figure}

\begin{figure}[h!]
    \centering
    \includegraphics[width=0.4\textwidth]{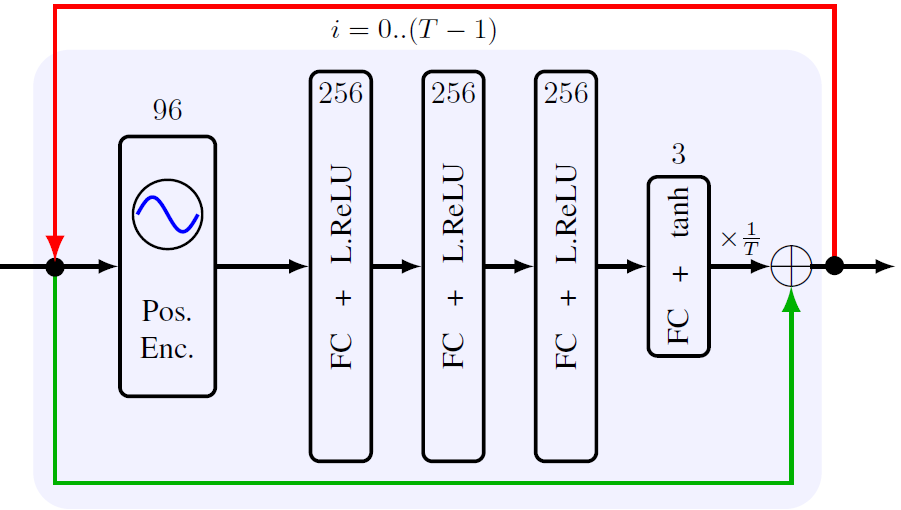}\vspace{-0.1in}
    \caption{\small Our Shape-Net, which is built upon an MLP with a global residual link and a feedback loop, forming a recurrent ResNet. It takes a 3D vector as input, and outputs a 3D vector. We iteratively call the ResNet block $T$ times, and re-scale the MLP output by $(1/T)$ after each iteration.}
    \label{fig:shape-Net}
    \vspace{-0.3cm}
\end{figure}
\subsection{Network Architecture}
In the preceding section, we have reduced the task of joint shape and {\small{svBRDF}} recovery to the task of finding two functional mappings, $\Phi:\vS\mapsto \vM$ and $\Psi:\vS\mapsto \mathbb{R}^5$.  

Here we will scribe our neural-network based solution.  Specifically, our method consists of two small networks: a Shape-Net to learn the spherical embedding diffeomorphism~$\Phi$, and a {\small{BRDF}}-Net to predict the {\small{svBRDF}} map~$\Psi$.

The overall architecture of our method is illustrated in Fig- \ref{fig:neural_networks_concept}. The left half of the figure depicts the Shape-Net and {\small{BRDF}}-Net.  The Shape-Net and the BRDF-Net are trained together by a differentiable renderer shown in the right half of Fig- \ref{fig:neural_networks_concept}.

Fig-\ref{fig:shape-Net} reveals the internal structure Shape-Net. From the figure, one can see that it is based on the MLP as the backbone, which maps a 3D vector to a 3D  vector. However, it has two distinctive features: (1) the MLP contains a global residual connection (colored in green in the figure), making it a single Residual block \cite{he2016deep}; (2) more importantly, a global feedback loop is added end-to-end to the above residual block, making the Shape-Net a recurrent ResNet. During training, the same ResNet block will be called iteratively for a number of $T$ times ($T=20$ in our experiments) before outputting the final output $\vx\in\mathbb{R}^3$.  In the next Section (Sec-\ref{sec:proof}), we will provide a formal justification as to why we design the Shape-Net this particular way, \ie a ResNet with a global feedback loop (\ie a `ring' connection).  In particular, We will prove that this {\em feedback loop} plays an essential role to our method, in the sense that it guarantees the Shape-Net always finds a valid {\em diffeomorphic} shape parametrization.  For now, let us simply allude to this as follows: this recurrent Residual Shape-Net in effect solves a time-continuous {\em diffeomorphism-defining} dynamic ODE system approximately up to certain time discretization. The number of iterations, $T$, corresponds to the discretized time steps.  One technical detail is, inspired by recent work \cite{mildenhall2020nerf,rahaman2019spectral}, we apply a positional encoding layer to the input vector to the network, aiming to better capture high-frequency details of the signal. Our {\small{BRDF}}-Net is designed as a regular 6-layer MLP (with positional encoding), which takes a 3D position on the unit sphere as input, and outputs a 5D {\small{BRDF}} parameters at that location (see the Appendix). Both networks are small and lightly parameterized. Shape-Net has only 771 neurons, while BRDF-Net has 1,285 nodes.

\subsection{The Training Process} 
Although the new method proposed in this paper makes use of neural networks, we still follow the traditional processing pipeline of Multi-View stereo based 3D-Reconstruction \cite{jensen2014large}.  Given multi-view images of an object as input, we jointly train the two neural nets (namely, the Shape-Net and BRDF-Net) to parameterize the unknown shape and unknown svBRDF.  We cast the 3D reconstruction problem as an optimization, following the general paradigm of {\em ``vision as inverse graphics"} \cite{romaszko2017vision,yu2019inverserendernet,tung2017adversarial,kulkarni2015deep}.

We use a differentiable renderer to account for the multi-view geometric and photometric constraints provided by the input images. Despite the objective function itself remains highly non-convex and extremely difficult to optimize with traditional optimization algorithms, we show in this paper that in all experiments our networks always converge to high quality solution that is globally optimal.

The overall loss function that we use to train the networks is a weighted sum of the rendering loss and a regularization term for shape deformation, \ie, 
\begin{equation}
L=L^{rgb}+\lambda L^{reg}.
\end{equation}
We set $\lambda =0.01$ empirically in all our experiments. The regularization term will be explained in Section-\ref{sec:proof}.

\subsection{The Image Rendering Loss} 

Our Shape-Net and {\small{BRDF}}-Net are trained jointly, via a differentiable renderer, condition on the multi-view input images.  Here we adopt the {\em soft rasterizer} \cite{liu2019soft} as the renderer for its simplicity. Other options are applicable as well (\eg\cite{petersen2019pix2vex,chen2019learning}).  We use the following physically-based shading equation:
\begin{equation}
    I^\text{prd}({\texttt{P}\vx}) = \max(0,\vn_\vx\cdot\vi_\vx) B(\vn_\vx,\vi_\vx,\vo_\vx;\theta_\vx)/{d_\vx^2},
\end{equation}where $\texttt{P}$ denotes perspective camera projection, $d_\vx$ is the distance from the point light source to $\vx$ on the object surface (or simply $d_\vx=1$ for distant/parallel light sources), and $B(\cdot;\theta_\vx)$ is its {\small{BRDF}} evaluated at this point. Without loss of generality, the light source intensity, camera response curve are subsumed in $\rho^{r,g,b}$ and $\rho^{\gamma}$.

By comparing the predicted (rendered) image $I_k^\text{prd}$ with the observed image $I_k^{obs}$ at camera view $k$, we get the rendering loss: $L^\text{rgb} = \sum_k\|I_k^\text{prd}-I_k^\text{obs}\|^2$.

When foreground object masks are available, one may use them to constrain the object's outlines (cf. \cite{liu2019soft}). However, in experiments we found while such step led to a faster convergence, the difference in the final shape is negligible.

Although the Shape-Net is able to approximate continuous maps (of $\Phi$), which suggests that the obtained object shape is a continuous 2-manifold surface (of infinite resolution), most off-the-shelf graphics renderers are however designed for discretized meshes. In order to employ these existing renderers, we sample points on the unit sphere and form a triangulated mesh structure. These sampled points are fed into the networks to compute its 3D position and BRDF. We pass this information to the renderer to render a predicted image. In our experiments, we use randomly rotated icosphere (\ie sub-divided icosahedron) for this purpose, collecting all vertices on the icosphere as one batch during training.  During testing time, one can generate the object shape and svBRDF up to an infinite resolution, and is not restricted to any particular mesh structure used during training. Alternatively, one could apply the implicit surface rendering technique (such as \cite{jiang2020sdfdiff,niemeyer2020differentiable}) to render continuous surfaces; however, the computational burden is significantly higher than that of mesh-based renderers. 

\section{Theory: why does it work?}\label{sec:proof}

Our Shape-Net has the task of modelling the shape of a genus-zero
embedded surface in $\R^3$, in other words, any embedded surface
$\vM$ topologically equivalent (homeomorphic) to a sphere $\vS$.

It will be assumed that the target surface $\vM$ is embedded in such a way that there is a {\em smooth flow} on $\R^3$ that takes $\vS$ at time $t=0$ to target surface
$\vM$ at time $t=1$. This assumption will be examined in more detail in the Appendix.  It is equivalent to saying that there
exists a smooth vector field $V$ defined on $\R^3$, and for every point $s \in \vS$ a curve $\gamma_s(t)$ in $\R^3$, defined for 
$t \in [0, 1]$, satisfying
{\small 
\begin{enumerate}
\item $\gamma_s(0) = s$;
~~~$\gamma_s(1) \in \vM$;
\item $\gamma_s'(t) = V\big(\gamma_s(t)\big)$.
\end{enumerate} 
}
The curve $\gamma_s$ is said to be an integration curve of the vector
field $V$, with initial point $s$.
The reader is referred to \cite{lee2013smooth},
chapter 9 for a detailed treatment of flows and vector fields
on manifolds.  They will also be considered in more detail in the Appendix.
In the present case, we are dealing
with vector fields on $\R^3$, and so a smooth vector field
is simply a smooth function $V: \R^3 \rightarrow \R^3$, assigning
a vector $V(x)$ in $\R^3$ to every point $x \in \R^3$. 

Given such a vector field, one may define a mapping
\[
\Phi(x, t) : \R^3\times I \rightarrow \R^3
\]
where $I$ is some interval in $\R$,
defined so that $\Phi(x, t)$ is the point 
reached by integrating along the vector field, starting
from time $0$ at point $x$ and integrating until time $t$.
In other words,
\begin{equation}
\Phi(x,t)=\gamma_x(t)=x+\int_0^t\gamma_x'(s)\,ds, \text{for}~ t\in I.
\end{equation} 

In general, define
$\vM_t$ to be the subset of points $x\in \R^3$ such that 
$\Phi(x, t)$ is
defined, and denote by $\Phi_t: \vM_t \rightarrow \R^3$
the mapping defined by $\Phi_t(x) = \Phi(x, t)$. Then
the {\em Fundamental Theorem of Flows} (see 
\cite{lee2013smooth}, theorem
9.12) states (in part) that for any fixed $t$, the mapping
$\Phi_t : \vM_t \rightarrow \R^3$ is a diffeomorphism onto its range, $\Phi_t(\vM_t)$.

In this description it will be assumed that the vector
field is integrable from time $t=0$ until $t=1$, since 
the vector fields that are constructed by our method will
have this property by construction. 

Therefore $\Phi_t$ is
defined for all $x$, and is a diffeomorphism of $\R^3 \rightarrow \R^3$ for all
$t \in [0, 1]$.

\subsection{Learning diffeomorphism via Shape-Net}

Our task, therefore, reduces to learning a diffeomorphism taking
$\vS$ to a desired embedded surface $\vM$.  This is solved by finding a smooth
velocity field $V$ on $\R^3$, such that by integrating 
$V$ from $t=0$ to $t=1$ a diffeomorphism $\Phi_1(x)$
from $\R^3$ to $\R^3$ will be defined. This diffeomorphism
restricts to a map of the 
surface $\vS$, mapping it to a surface $\Phi_1(\vS) = {\vM}$ representing the shape of the object being reconstructed, and at the same time minimizing the image rendering loss $L^{rgb}$.

The desired velocity field is modelled by the MLP in Shape-Net,
and integration of the vector field will be carried out by a sequence of $T$
steps of length $1/T$, known as Euler steps.  This is essentially
the Euler method of solving the diffeomorphism-defining ODE as follows.
\begin{equation}
\frac{d\Phi(x,t)}{dt} = V(\gamma_x(t)).
\end{equation} 
Therefore, recursively running the MLP for $T$ times, constituting the
feedback residual loop in {\em Shape-Net}, maps $\vS$ to the desired surface.

To improve the numerical integrability of above ODE, a regularization loss is placed on the MLP to make it sufficiently smooth: \begin{math}
    L^{reg} = \sum_{s,t}\|(\mathbf{I}-\alpha\Delta)\circ V\big(\gamma_s(t)\big)\|^2,
\end{math} where $\mathbf{I}$ and $\Delta$ are identity and Laplacian operators, and $\alpha$ a weight parameter. 

\begin{figure}[h!]
    \centering
    \includegraphics[width=0.5\textwidth,trim={ 0cm 0cm 0cm 0cm}, clip]{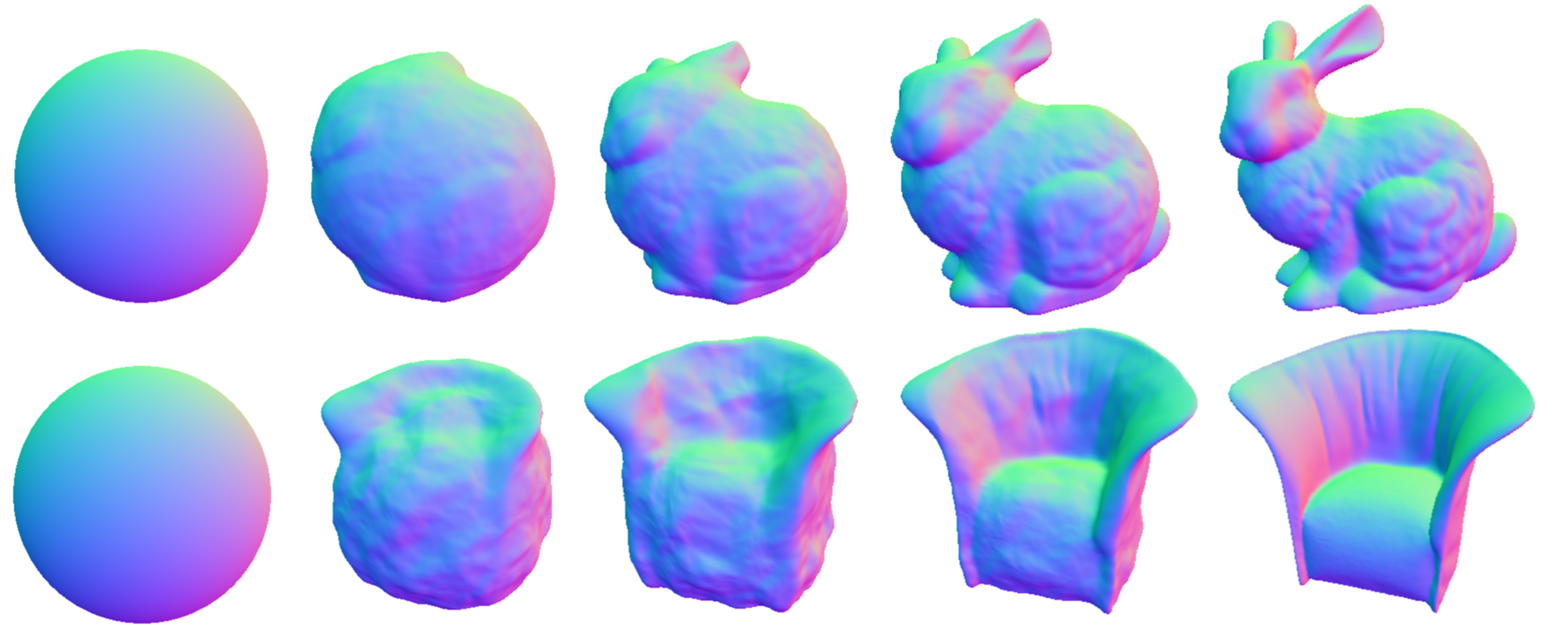}
    \caption{\small We show how the recovered shape evolves over time at t=0,1/4,1/2,3/4,1. Shapes are color-coded with surface normals for better visualization.}
    \label{fig:deform}
    \vspace{-0.3cm}
\end{figure}

With sufficiently large $T$, the Shape-Net solves a diffeomorphism $\Phi$ by integration. Fig-\ref{fig:deform} shows how the recovered shape evolves gracefully over $T$ steps. Surface details began to emerge early on in the iterations.
In contrast, neural networks by themselves do not generate a diffeomorphic mapping in general. To demonstrate this, we conducted an ablation study in the Appendix which shows that a plain ResNet without the ring connection suffers from surface self-intersection and back-facing, while our Shape-Net generates a surface free of these artefacts. Alternatively, one could also use the Neural-ODE solver \cite{chen2018neural} to solve the integration curves. This solver however requires solving two ODEs -- one forward and one backward, hence is much more time/space consuming (cf. \cite{salman2018deep,chen2018neural,rousseau2019residual}).

To conclude, we emphasize that the global feedback loop (the `ring connection') plays an essential role to our method--by which we are able to jointly solve both shape and materials --hence the paper's title of {~\em``one ring to rule them all" \cite{LOR}}.

\input{ICCV21_Experiments}
\input{ICCV21_RelatedWorks} 
\section{Closing Remark}
We have developed a novel neural-networks based solution which solves a challenging open problem of joint multi-view shape and reflectance recovery under fewer constraints. Both the camera and light source are allowed to roam freely, while the object can have unknown, arbitrary, and spatially-varying reflectance.  Our core contribution is a small recurrent ResNet block, which implements a provable diffeomorphic shape parametrization, offering a convenient parameterization to a 3D shape even before it is reconstructed.

Our method achieves state-of-the-art performance, and opens up new opportunities for novel applications and future research. For example, 1. the current {\em per object} learning may be able to generalize to {\em per category} learning; 2. Our current method is still restricted to a darkroom environment due to the use of active light. In the future we will explore how to extend this work to natural environment lighting conditions; 3. our current rendering equation ignores shadows and inter-reflections; both shall be tackled in our future work.

Moreover, from a theoretical point of view, it is desirable to understand {\em why} and {\em under what condition} a simple neural network may be able to reduce a non-convex optimization problem to a much simpler one, readily solvable by the standard stochastic gradient descent. Our method is easy to implement, works robustly, and reaches superior performance in terms of shape/BRDF accuracy; we hope this paper will inspire further research. Our codes and models will be released to the community. 
\vspace{-0.2in}\paragraph{Acknowledgement.} {\small We wish to thank Prof Boxin Shi for sharing the DiLiGent-MV dataset. We thank Dr Art Subpa-asa for drawing the artistic diagrams for the paper. ZC and HL are funded by the Australia Research Council (ARC) via a Discovery Grant project (DP190102261).}
\vspace{-0.1in}
{\small
\bibliographystyle{ieee}
\bibliography{egbib}
}
\input{ICCV21_SupplementMaterial}
\end{document}

%% file: ICCV21_abstract.tex
This paper proposes a simple method which solves an
open problem of multi-view 3D-Reconstruction for objects
with unknown and generic surface materials, imaged
by a freely moving camera and a freely moving point
light source. The object can have arbitrary (\eg non-
Lambertian), spatially-varying (or everywhere different)
surface reflectances (svBRDF). Our solution consists of two
small-sized neural networks (dubbed the ‘Shape-Net’ and
‘BRDF-Net’), each having about 1,000 neurons, used to parameterize
the unknown shape and unknown svBRDF, respectively.
Key to our method is a special network design
(namely, a ResNet with a global feedback or `ring' connection), which has a provable guarantee for finding a valid diffeomorphic shape parameterization. Despite the underlying problem is highly non-convex hence impractical
to solve by traditional optimization techniques, our method
converges reliably to high quality solutions, even without initialization.
Extensive experiments demonstrate the superiority of our method, and it naturally enables a wide range of special-effect applications including novel-view-synthesis, relighting, material retouching, and shape exchange without additional coding effort. We encourage the reader to view our demo video for better visualizations.

%% file: ICCV21_Experiments.tex
\section{Experiments}
We validate the proposed method on synthetic and real-world objects under different camera-light configurations, and compare with existing methods. Networks are trained with Adam optimizer \cite{kingma2014adam} with $lr=0.0001$ for 2K epochs. Timing etc are reported in the Appendix.  Relighting results for shiny objects are best viewed as videos, so we encourage the reader to view our accompanying video for convincing visualizations.

\begin{figure*}
    \centering
    \includegraphics[width=1\textwidth]{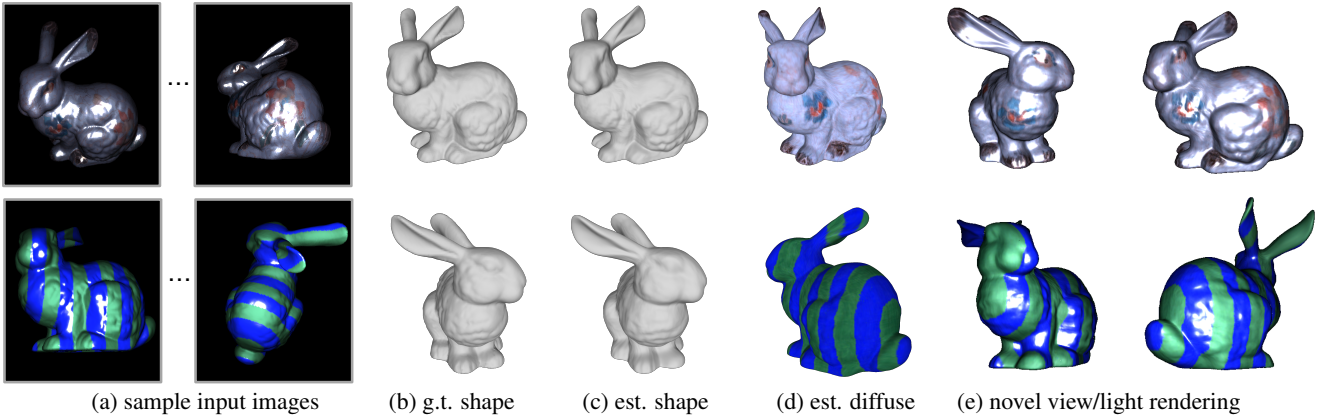}
    \begin{subfigure}[b]{0.23\textwidth}
    \centering\caption{sample input images}
    \end{subfigure}
    \begin{subfigure}[b]{0.135\textwidth}
    \centering\caption{g.t. shape}
    \end{subfigure}
    \begin{subfigure}[b]{0.15\textwidth}
    \centering\caption{est. shape}
    \end{subfigure}
    \begin{subfigure}[b]{0.045\textwidth}
    \end{subfigure}
    \begin{subfigure}[b]{0.135\textwidth}
    \centering\caption{est. diffuse}
    \end{subfigure}
    \begin{subfigure}[b]{0.24\textwidth}
    \centering\caption{novel view/light rendering}
    \end{subfigure}
    \vspace{-0.2cm}
    \caption{\small Two synthetic bunny models reconstructed by our method, the top one coated with evenly glossy materials with different albedo/texture, and the bottom one with two different materials (shinny and dull) in alternation.  Our method robustly recover the surface shape, albedo and specularities for novel view/re-lighting rendering. {\bf Better viewed on screen and in the companion demo video.}}
    \label{fig:bunnies}
    \vspace{-0.2cm}
\end{figure*}

\begin{figure*}
    \centering
    \includegraphics[width=0.9\textwidth]{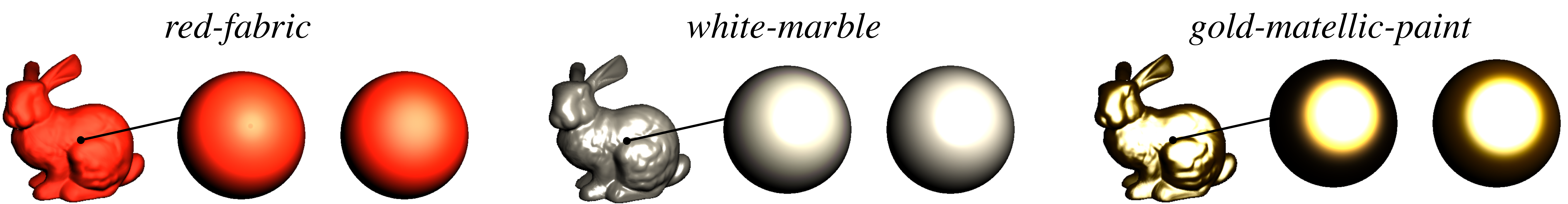}
    \caption{\small BRDF recovery by our method. Here we show the re-rendered shape recovery (left), predicted BRDF at a surface point (middle), and the corresponding ground-truth BRDF (right).
    \label{fig:BRDF}}
    \vspace{-0.5cm}
\end{figure*}

\vspace{-0.1in}\subsection{Synthetic objects}
We render multi-view images of 7 objects (\textit{Bunny}, \textit{Girl}, \textit{Head}, \textit{Pig} \textit{Sofa}, \textit{Teapot}, and \textit{Tool}) under a perspective camera (60 degrees field of view) and near-field light source. We simulate a collocated lighting configuration, where the light source is rigidly attached to camera to ease light calibration and eliminate shadows. For realistic rendering, target objects are coated with real-world materials from MERL dataset \cite{Matusik2003A}, and the rendered images are digitized to standard dynamic range of 8-bit instead of the more demanding HDR required by previous methods. We render 50 images from 50 random viewpoints.

Our synthetic dataset includes both textured and texture-less objects with various degrees of glossiness and overexposure, all imaged from non-overlapping viewpoints and lighting conditions. We note such generic setting differs from conventional methods in Shape-from-Motion and Photometric Stereo categories and could prove to be particularly challenging to them.

\begin{figure}
    \centering
    \includegraphics[width=0.45\textwidth]{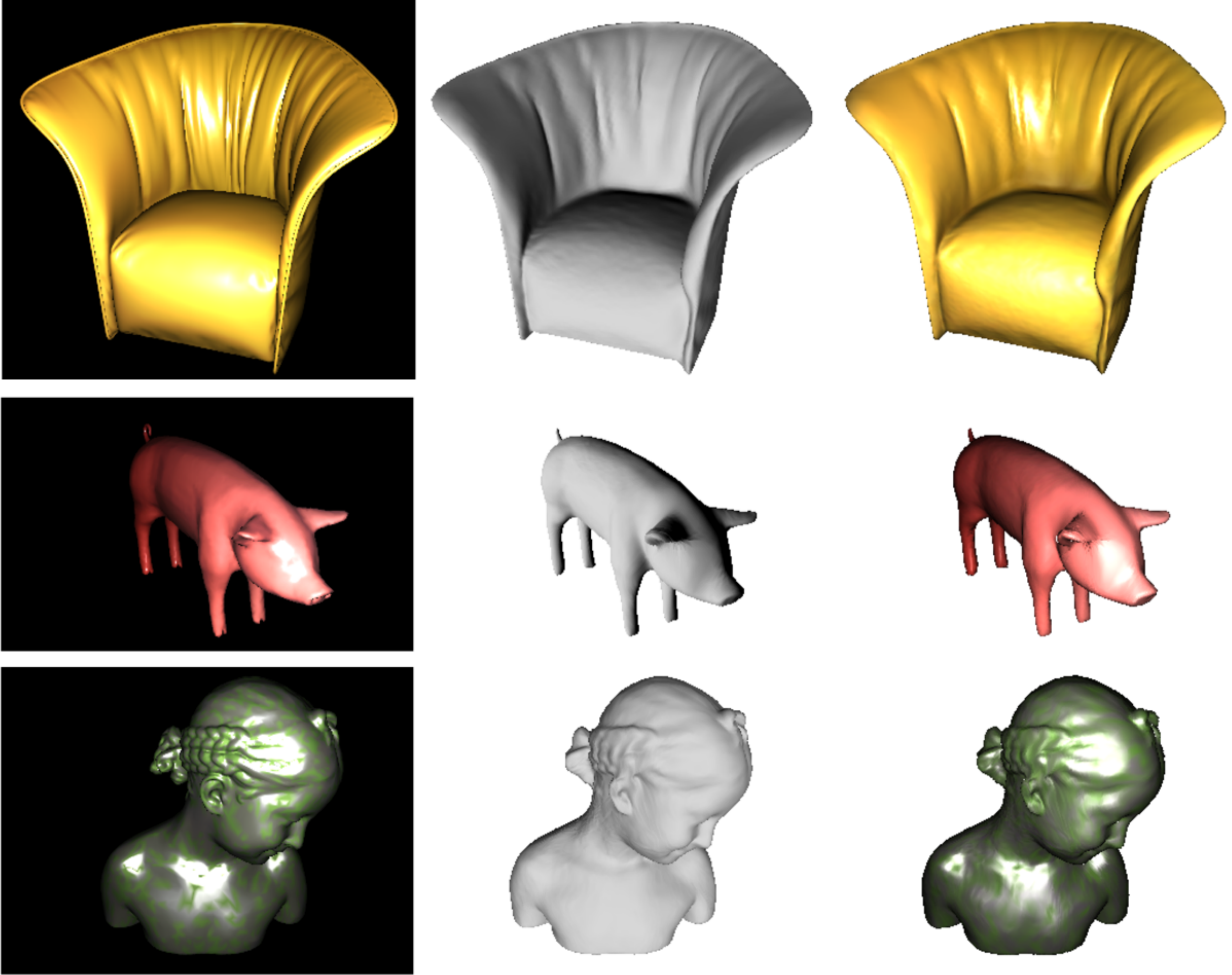}
    
    \begin{subfigure}[b]{0.14\textwidth}
    \centering\caption{\scriptsize input}
    \end{subfigure}
    \begin{subfigure}[b]{0.14\textwidth}
    \centering\caption{\scriptsize est. shape}
    \end{subfigure}
    \begin{subfigure}[b]{0.14\textwidth}
    \centering\caption{\scriptsize re-render}
    \end{subfigure}
    \vspace{-0.5cm}
    \caption{\small Example shape and svBRDF recovery in synthetic experiments.}
    \vspace{-0.5cm}
    \label{fig:synthetic}
\end{figure}

We evaluate the accuracy of shape reconstruction for each input view. Table~\ref{tab:synthetic} shows the mean and median errors of recovered normal and depth in degrees and in percentages of object dimension respectively. On the other hand, direct {\small svBRDF} evaluation is difficult due to the fact that high specular peaks are saturated and hence cannot be exactly recovered. Instead, to validate the accuracy of reflectance estimation, we render recovered objects from 50 novel viewpoints/lights and compare the predicted images with corresponding ground truths. We tabulate the {\small PSNR} metric for both input and novel view renderings in Table~\ref{tab:synthetic}. The results suggest our {\small svBRDF} estimations generalize well to novel view/light angles. Figure~\ref{fig:synthetic} illustrates the qualitative results on several synthetic objects. 

\vspace{-0.1in}
\paragraph{Robustness to specularities.} We simulate \textit{Bunny} with three uniform MERL materials of increasing glossiness. Our aim is to validate the robustness our model to varying specularities. Fig-\ref{fig:BRDF} shows the recovered reflectances match the ground-truth faithfully, and the shape is well restored regardless of materials. To further examine robustness to spatially varying glossiness, we render a bunny model coated in two distinct materials, one diffuse and one specular interlaced in a stripe pattern.  Figure~\ref{fig:bunnies} illustrates the reconstruction results, compared to the plain bunny of evenly glossy material.

{\small \begin{table}[h]
    \centering
    \footnotesize
    \begin{tabular}{l|c*5{|c}}
     & \multicolumn{2}{c|}{Normal error(deg)} & \multicolumn{2}{c|}{Depth error (\%)} & \multicolumn{2}{c}{PSNR (dB)} \\
     &     \footnotesize{ Mean}     &   \footnotesize{  Median}     &  \footnotesize{   Mean}      &    \footnotesize{Median }     & Input & \footnotesize{Novel}\\\hline
    Bunny & 3.61 & 2.21 & 0.18 & 0.09 &  33.9 & 33.5 \\ \hline
    Girl & 4.81 & 1.84 & 0.40 & 0.14 & 31.1 & 30.2 \\ \hline
    Head & 2.45 & 1.50 & 0.16 & 0.09 & 32.0 & 29.5 \\ \hline
    Pig & 4.10 & 1.97 & 0.22 & 0.11 & 36.5 & 37.4 \\ \hline
    Sofa 
          & 5.44 & 2.12 & 0.55 & 0.26 & 30.0 & 29.9\\ \hline
    Teapot & 5.75 & 2.55 & 0.30 & 0.11 & 29.8 & 27.6 \\ \hline
    Tool & 3.00 & 1.18 & 0.24 & 0.14 & 33.1 & 31.6 \\ \hline
    \end{tabular}
    \centering
    \caption{\footnotesize Errors in surface normal and depth, and PSNR for image re-rendering on synthetic objects. Depth errors are measured relative to the object's bounding box size. A total of 50 input views were used for training, and another 50 real images held out for validation.}
    \label{tab:synthetic}
    \vspace{-0.3cm}
\end{table}}

\begin{figure*}
    \centering
    \includegraphics[width=1.0\textwidth]{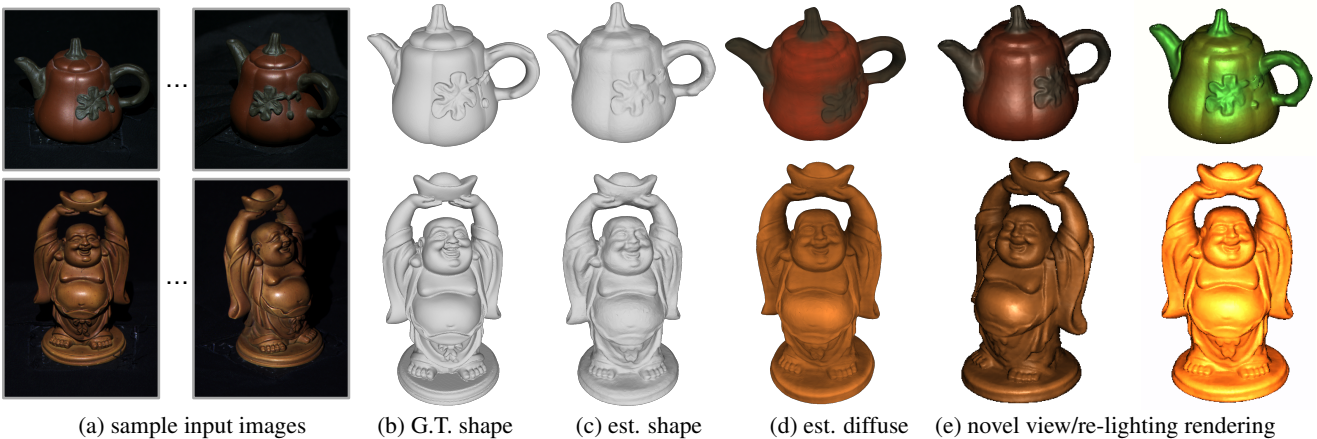}
    \begin{subfigure}[b]{0.24\textwidth}
    \centering\caption{sample input images}
    \end{subfigure}
    \begin{subfigure}[b]{0.135\textwidth}
    \centering\caption{G.T. shape}
    \end{subfigure}
    \begin{subfigure}[b]{0.15\textwidth}
    \centering\caption{est. shape}
    \end{subfigure}
    \begin{subfigure}[b]{0.045\textwidth}
    \end{subfigure}
    \begin{subfigure}[b]{0.135\textwidth}
    \centering\caption{est. diffuse}
    \end{subfigure}
    \begin{subfigure}[b]{0.26\textwidth}
    \centering\caption{novel view/re-lighting rendering}
    \end{subfigure}
    \vspace{-0.5cm}
    \caption{\small Results on DiLiGenT-MV objects \textbf{Pot2} (top) and \textbf{Buddha} (bottom) \cite{li2020multi}. We reconstruct shape and svBRDF for realistic novel view/re-lighting rendering. {\bf Better viewed on screen.}}
    \label{fig:diligent_big}
    \vspace{-0.4cm}
\end{figure*}
\vspace{-0.1in}
\subsection{Real-world experiments and comparisons} Few existing 3D reconstruction methods address the same challenging problem setting that we aim to solve, \ie freely moving camera and unstructured light source over arbitrary unknown svBRDF. Due to limited available implementations and datasets, we compare our method with Park~\etal~\cite{park2016robust} as an MVPS baseline and {\small ColMAP} \cite{schoenberger2016sfm} as the state-of-art SfM approach, for qualitative comparisons. We did not compare with the native DiliGent-MV solver as it requires specialized imaging devices (\ie concentric light sources) and manual correspondence matching \cite{li2020multi}. 

\begin{figure}
    \centering
    \tiny
    \includegraphics[width=0.48\textwidth]{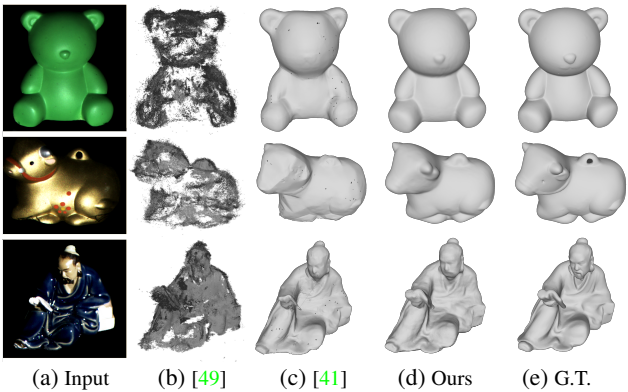}
    \begin{subfigure}[b]{0.09\textwidth}
    \centering\caption{\footnotesize Input}
    \end{subfigure}
    \begin{subfigure}[b]{0.09\textwidth}
    \centering\caption{\footnotesize \cite{schoenberger2016sfm}}
    \end{subfigure}
    \begin{subfigure}[b]{0.09\textwidth}
    \centering\caption{\footnotesize \cite{park2016robust}}
    \end{subfigure}
    \begin{subfigure}[b]{0.09\textwidth}
    \centering\caption{\footnotesize Ours}
    \end{subfigure}
    \begin{subfigure}[b]{0.09\textwidth}
    \centering\caption{\footnotesize G.T.}
    \end{subfigure}
    \vspace{-0.5cm}
    \caption{\small Comparison with existing method on real-world objects, \textbf{Bear} (top), \textbf{Cow} (middle) and \textbf{Reading} (bottom), from DiLiGenT-MV dataset. Note that method \cite{park2016robust} used quality initial meshes, while ours needs no initialization. Despite this, quantitatively our method outperforms both competing methods (see Table-\ref{tab:diligent-mv}). {\bf Better viewed on screen with zoom-in.}\label{fig:diligent}}
    \vspace{-0.3cm}
\end{figure}

Figure~\ref{fig:diligent} illustrates three reconstructed objects (bear, cow and reading) by different methods. {\small ColMAP} \cite{schoenberger2016sfm} yields noisy sparse point clouds due to large texture-less and specular regions. Also, similar to most multi-view photometric approaches, \cite{park2016robust} used a high-quality initial mesh to start their algorithm, which was obtained from an external SFM pipeline \cite{galliani2015massively} aided by human interventions \cite{li2020multi}. In contrast, our method is initialized from random weights and outperforms both significantly. Our normal and depth errors on real images are consistent with that on synthetic images.

\vspace{-0.1in} \paragraph{Higher genus reconstruction.} We also tested two genus-one shapes in the DiLiGent-MV dataset, Pot2 and Buddha, as shown in Figure~\ref{fig:diligent_big}. To accommodate these higher-genus objects, we use their visual hull reconstruction as the embedding space as opposed to the unit sphere. We visualize the reconstructions under novel viewpoint and new lighting conditions. As can be seen, the re-rendered images preserve both geometric details and surface specularities.

\vspace{-0.2in}\paragraph{Quantitative Comparison.} Table~\ref{tab:diligent-mv} summarizes quantitative evaluations on surface normal and depth. We include two state-of-art photometric stereo baselines Zheng~\etal\cite{zheng2019numerical} and Enomoto~\etal\cite{enomoto2020photometric}, as well as Park~\etal~\cite{park2016robust}. Our method outperforms on both metrics by a large margin, and achieves sub-millimeter accuracy on all objects. Compared with photometric stereo methods that recover normal map from a static viewpoint, our normal errors are much smaller on average. 

\begin{figure}
    \centering
    \includegraphics[width=0.45\textwidth]{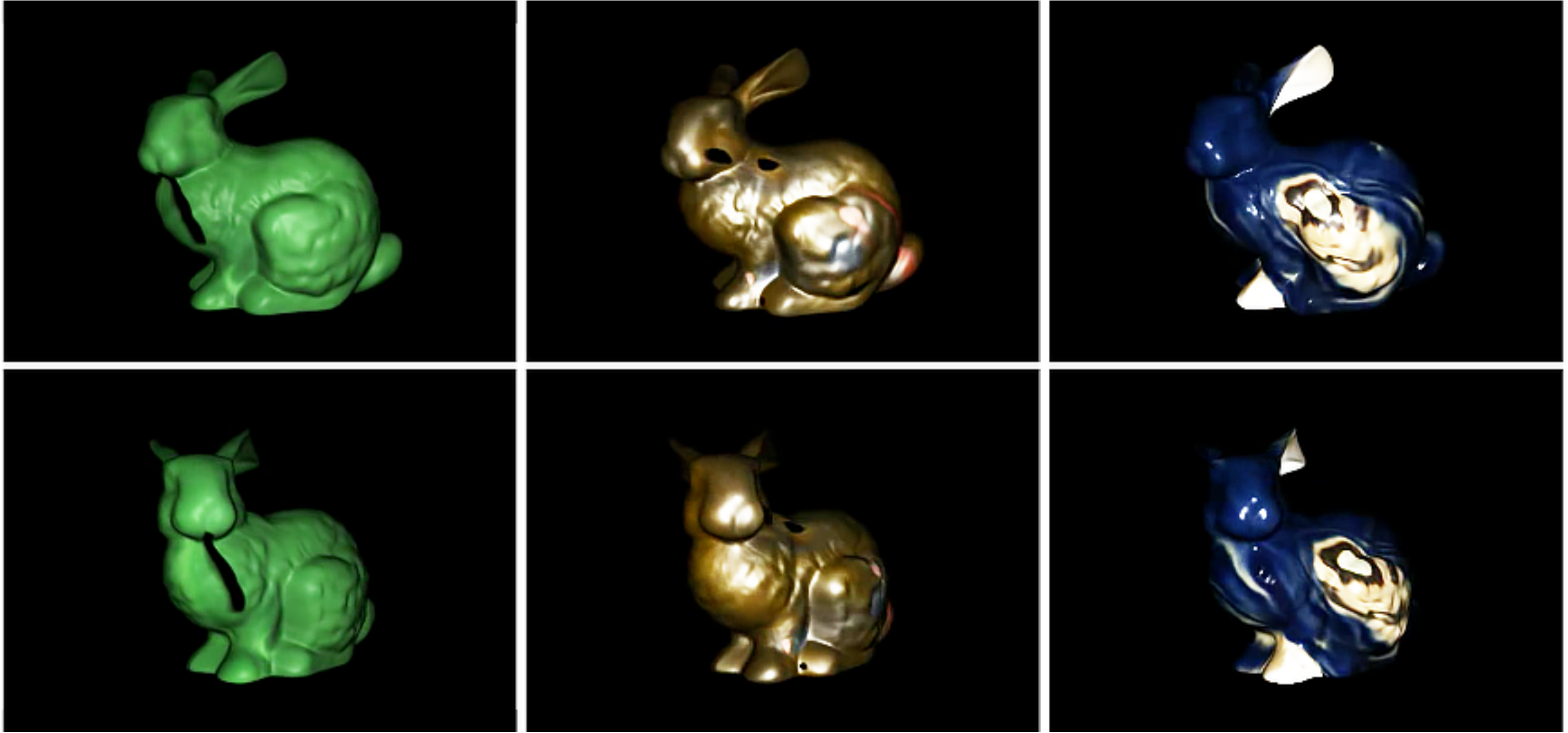}
    
    \begin{subfigure}[b]{0.13\textwidth}
    \centering\caption{\scriptsize Bunny+Bear}
    \end{subfigure}
    \begin{subfigure}[b]{0.13\textwidth}
    \centering\caption{\scriptsize Bunny+Cow}
    \end{subfigure}
    \begin{subfigure}[b]{0.13\textwidth}
    \centering\caption{\scriptsize Bunny+Reading}
    \end{subfigure}
    \vspace{-0.2cm}
    \caption{\small Our method naturally allows for easy material exchange, simply by swapping the recovered svBRDF maps.  Shown here are the re-rendered Bunny with the materials (svBRDF) of the Bear, Cow and Reading.}
    \label{fig:re-material}
    \vspace{-0.3cm}
\end{figure}

\vspace{-0.2in}\paragraph{Materials Exchange.} Since our method recovers shape and {\small svBRDF} embeddings separately from two networks, one can easily swap the svBRDF maps from one object to other if the two are embedded in the same domain. Here we show material editing by our method by combining the Shape-Net trained on synthetic bunny with BRDF-Net trained on DiLiGenT-MV objects. The results are illustrated in Fig~\ref{fig:re-material}.

\vspace{-0.1in}\paragraph{Camera/light calibration refinement.} In all our previous experiments, we use pre-calibrated camera poses and light positions. In fact, within the same framework of our neural solver, we are able to refine the camera and light parameters as well (see Supp. Material for details).  

{\footnotesize
\begin{table}[h!]
    \centering
     \footnotesize
    \begin{tabular}{c*6{|c}}
                                   &  & Bear & Buddha& Cow & Pot2 & Reading \\\hline
    \multirow{4}{*}{Normal}  & Ours& \textbf{4.42} & 12.08 & \textbf{4.21} & \textbf{6.63} & \textbf{7.61}  \\ \cline{2-7}
                                 & \cite{park2016robust} & 12.52 & 13.71 & 10.64 & 14.59 & 11.45   \\\cline{2-7}
                                 & \cite{zheng2019numerical} & 4.65 & \textbf{9.14} &  15.85 & 8.09 & 12.77   \\\cline{2-7}
                                 & \cite{enomoto2020photometric} & 6.38 & 13.69 &  7.80 & 7.26 & 15.49   \\\hline
                      
    \multirow{2}{*}{Depth} & Ours& \textbf{0.75} & \textbf{0.67} & \textbf{0.77} & \textbf{0.58} & \textbf{0.58}  \\ \cline{2-7}
                      &\cite{park2016robust} &  1.89 & 1.28 & 0.85 & 3.03 & 1.24   \\\hline
    \end{tabular}
    \caption{\footnotesize Shape recovery errors in Normal (in `degree') and Depth (in `mm') on real images from the DiLiGenT-MV benchmark.} 
    \label{tab:diligent-mv}
\vspace{-0.3cm}
\end{table}
}

%% file: ICCV21_RelatedWorks.tex
\section{Other Related Work}
\paragraph{Traditional Multi-view/Photometric SfM.}
Traditional multi-view Structure-from-Motion (SfM) methods (\eg \cite{agarwal2011building,schoenberger2016sfm,vijayanarasimhan2017sfm, PMVS}) cannot handle highly specular non-Lambertian surfaces due to the difficulty in establishing feature correspondences.  Traditional physically-based 3D-reconstruction methods (such as photometric stereo), on the other hand, often require special instrumented camera equipment in a lab/studio environment. Moreover, they often involve solving a complex optimization problem demanding a very good initialization (\eg \cite{nam2018practical,schmitt2020joint,li2020multi,Zhou2013multi}). To circumvent these issues, many previous methods assume diffuse/Lambertian reflectance \cite{higo2009hand,vlasic2009dynamic,wu2010fusing,delaunoy2014photometric,park2016robust,oxholm2012shape} which overly simplifies the task.  Nam \etal ~\cite{nam2018practical} use a collocated camera-light scanner for shape reconstruction.  Cheng and Li \etal \cite{VincentCVPR21} develop a new method also for the collocated setting, which avoids shape initialization by using a randomized PatchMatch optimization algorithm. Logothetis \etal \cite{logothetis2019differential} propose a volumetric parameterization under Lambertian assumption. There are methods which resort to RGB-D depth camera for shape initialization  \cite{maier2017intrinsic3d,bylow2019combining,schmitt2020joint,ha2020progressive}.  While these methods achieve decent results, many of them rely on careful engineering and hand-crafted priors on shape and reflectance through a tailored optimizer under restrictive imaging settings. 
\vspace{-0.2in}\paragraph{Deep learning 3D Reconstruction.}  Deep learning has been used for 3D object reconstruction (\eg, \cite{meshRCNN,vijayanarasimhan2017sfm,bi2020deep,niemeyer2020differentiable,marrnet,Wu_2020_CVPR,jiang2020sdfdiff}.) 
However, many of these methods, while giving good qualitative reconstruction, are lagging behind in terms of metric accuracy and visual fidelity. Several recent neural networks learn to predict a mesh from input images. Our method shares some similarity to \cite{wang20193dn,pixel2mesh++} in terms of driving a deformable initial shape towards the target shape, however, both the network designs and underlying theories are fundamentally different. Gupta \etal \cite{gupta2020neural} also formulates shape deformation as a flow using  Neural ODE\cite{chen2018neural}, but they require the ground truth shape for training by examples.  Neural differentiable rendering has been employed to regress simple meshes (\eg  \cite{petersen2019pix2vex,chen2019learning,liu2019soft,niemeyer2020differentiable,chang2015shapenet}), yet their generalization ability is limited by the training examples too. Neural models have been used for shape representation, either for surfaces (\eg \cite{atlasnet,foldingnet}) or for volumes (\eg \cite{park2019deepsdf,lombardi2019neural,mescheder2019occupancy}). When paired with a renderer, such network may be trained directly using images without 3D supervision \cite{bi2020deep,niemeyer2020differentiable}. Yariv \etal \cite{yariv2020multiview} proposed neural representations for objects' shape and appearance, but they cannot explicitly recover surface reflectance.  Despite our method is built upon neural networks, it is different from previous deep-learning based approaches mentioned above. Those learning-based methods often require pre-training on a large scale 3D dataset, and then predict a 3D shape from one or more input images. In contrast, our method follows the traditional optimization procedure, where the two small neural-nets are used to \textbf{re-parameterize} the shape and BRDFs, making otherwise difficult problem more easily to solve (simply by back-propagation and SGD). Our Shape-Net and BRDF-Net are trained on a {\em per object basis}. This is akin to NeRF \cite{mildenhall2020nerf}, in the sense that the networks learn to `remember' a particular object or scene. 

%% file: ICCV21_SupplementMaterial.tex
\title{Appendix}\author{}
\maketitle
\author{}

\appendix
\section{More Implementation Details}
\subsection{Positional encoding} 
We use the positional encoding layer to the input coordinates of both the Shape-Net and BRDF-Net in order to better predict high-frequency details.  The positional encoding layer is given as 
\begin{equation}
    PosEnc(\mathbf{x}) = \left[
    \begin{array}{c}cos(\omega\mathbf{x})\\ sin(\omega \mathbf{x})\end{array}\right],
\end{equation}
where $\omega$ takes values from $\{1,2,3, 4,..,16\}$. Different from the common practice which uses quadratic (\ie power of 2) frequency sweeping \ie $\omega=\{1,2,4,8,16,32,..\}$, we use a linear sweeping to achieve uniform coverage of frequencies.  The output of the $PosEnc$ layer is a 96-D vector ($=16\times 2\times 3$).  Our ablation study has confirmed the effectiveness the positional encoding layer, \ie  preserving more high frequency surface details. 

\subsection{The internal structure of the BRDF-Net}
\begin{figure}[h!] 
    \centering
\includegraphics[width=0.36\textwidth]{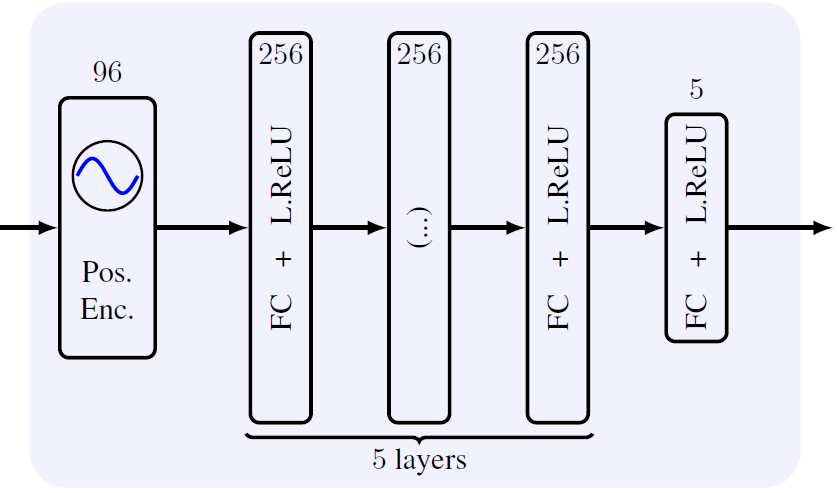}\vspace{-0.1in}
\caption{\small Our BRDF-Net, based on a simple MLP with positional encoding. It maps a 3D vector to a 5D BRDF (parametrized by Cook-Torrance coefficients).}
    \label{fig:BRDF-net}
\end{figure}

Our BRDF-Net (see fig-\ref{fig:BRDF-net}), which is a regular 7-layer MLP (with positional encoding), acts as a simple BRDF regressor which takes a position (on the unit sphere) as input, and predicts the BRDF of the corresponding point on the target object surface.  This is possible because our Shape-Net provides a one-to-one spherical parameterization of the surface.  

Our BRDF-Net is different from several recent works for deep-learning based BRDF estimation. The latter often directly take an image patch (\ie its visual appearance) as input and output (regress) the BRDF for that image patch.  In contrast our BRDF-Net takes 3D position as input and output the BRDF parameters at that point.  
 
\subsection{The Laplacian regularization term}

The Laplacian regularization term used in our loss function is defined as: 
\begin{equation}
    L^{reg} = \int_t\int_{\Omega}\|\mathcal{L} v({\Phi}(x,t))\|^2,
\end{equation}
where $\mathcal{L} = I- a\nabla^2$ with $a\in[0,1]$.  
The purpose of such a regularization term is to ensure that the flow field is sufficiently smooth, therefore improves the integrability of the diffeomorphism-defining ODE during numeric computation.

\subsection{Coarse to fine training (optional)}
One may optionally adjust the sampling rate to accelerate training. In our experiments, we found it helpful to first sample at a coarse level with fewer vertices, and/or render low resolution images. This significantly reduces the complexity of training, and allows networks to quickly converge to a coarse reconstruction. From this point on, we may gradually increase sampling rate and/or rendering resolution for a finer reconstruction. Such coarse to fine training can improve scalability of algorithm for handling large amount of high-resolution images. In practice we use a two-scale training routine: at the coarse level we sample 10k vertices and train for 2,000 epochs; we later increased vertices count to 41k and trained for another 4,000 epochs.

\subsection{Global regularization terms (optional)}  One could also optionally add other global priors to the loss function to better constrain the solution. For example, many natural or man-made shapes are piece-wise smooth, for which the TV-L1 (total variation L1) prior can be used.  Moreover, if the surface sv{\small{BRDF}} is spatially sparse, for which L1 or low rank regulation may be applied. Technically, all these priors can be effectively computed on a full batch of surface  points during network training.  However, in our experiments, we found that our network is already able to produce \eg piece-wise smooth geometry while preserving sharp geometry details, as well as sparse sv{\small{BRDF}}s without adding such additional prior terms.  This was somewhat surprising to us.  Without further investigation (doing which would beyond the scope of this paper), we attribute this to some inherent priors naturally enforced by the structure of deep networks.

\section{More Theoretic Analysis} 
\subsection{The existence of diffeomorphic embedding}
We examine the assumption that there exists a flow along
some vector field taking one surface $S_0$ to another surface
$S_1$ in $\R^3$, where $S_0$ and $S_1$ are embeddings of the 
standard sphere $S^2$.  

The Schoenflies theorem (or Jordan-Schoenflies theorem) states that
any simple closed curve (embedding of $S^1$) in the plane $\R^2$
separates the plane into two regions, the ``inside'' and the
``outside'', and that these two regions are homeomorphic to the 
inside and outside of the standard unit circle in the plane.

This theorem is not true in higher dimensions, without further
assumptions.  The most famous example in $\R^3$ is the 
Alexander Horned Sphere, an embedding of $S^2$ in
$\R^3$ that separates $\R^3$ into two parts, but the outside
is {\bf not} homeomorphic to the exterior of the standard
unit sphere.
This is an example of a so-called {\em wild embedding} of the sphere.
(Many images of the Alexander Horned Sphere exist on the internet.)
It is simple to extend the idea of the Alexander Horned Sphere
so that the interior region is not homeomorphic to the standard
unit ball, either.
Under these circumstances, it is therefore impossible that there should be a diffeomorphism,
or homeomorphism of $\R^3$ that takes the standard sphere to
an Alexander Horned Sphere.

However, the so-called Generalize Schoenflies Theorem (namely
the Schoenflies Theorem for higher dimensions) holds under the
additional assumption that the embedding of $S^{n-1}$ in $\R^n$
is a {\em collared embedding}\footnote{
Morton Brown. {\em A proof of the generalized schoenflies
theorem}.  Bulletin of the American Mathematical Society,
66(2):74–76, 1960.  This paper proved the Generalized Schoenflies Theorem for collared embeddings.  It is short and easily read (and entertaining) without requiring specialist knowledge of geometric topology.
}.
In other words there exists an embedding 
$\phi: S^{n-1} \times [-1,1] \rightarrow \R^n$ then 
the restriction of $\phi$ to $S^{n-1} \times \{0\}$ is a
so-called collared embedding of $S^{n-1}$ in $\R^n$.
Under these circumstances, the embedded sphere does separate
$\R^n$ into two parts homeomorphic to the partition of $\R^n$
by the standard unit sphere.

What this means in $3$-dimensions is that if $M$ is an
embedded sphere in $\R^3$, resulting from a collared embedding,
there there exists a homeomorphism of $\R^3$ that takes
the standard unit sphere $S_0$ to $M$.

Here, we shall call an embedding of $S^{n-1}$ in $\R^n$ that
results from a collared embedding a {\em tame} (as opposed
to {\em wild} embedding. 

In the smooth category (smoothly embedded spheres)
the Generalized Schoenflies Theorem holds in every 
dimension except possibly $n=4$, so we shall speak of
smooth embeddings rather than tame embeddings.

\paragraph{Deformations. }
We require more than that there exists a homeomorphism (or diffeomorphism) of
$\R^3$ that takes $S_0$ to $M$.  Our method requires that there
be a flow along some vector field on $\R^3$ that takes $S_0$ to $M$.
Such a flow will deform the surface $S_0$ smoothly to $M$
as time varies from $t=0$ to $t=1$.  

A theorem of Kirby, states that any orientation-preserving 
homeomorphism of $R^n$ is isotopic to
the identity mapping.  This means that given a homeomorphism
$h: \R^n \rightarrow \R^n$ there exists an isotopy, namely
a (continuous) mapping $\phi: \R^n \times [0, 1] \rightarrow \R^n$, 
such that 
\begin{enumerate}
\item $\phi(\cdot, 0)$ is the identity map on $\R^n$.
\item $\phi(\cdot, 1)$ is equal to $h$.
\item $\phi(\cdot, t)$ is a homeomorphism for all $t$.
\end{enumerate}
This expresses the fact that $R^n$ can be continuously deformed
to any homeomorphism.  This gives a continuous deformation 
of the standard sphere $S^{n-1}$ to any collared embedding
of the sphere.

The result of this and the Generalized Schoenflies Theorem
is that any two tame embedded spheres in $\R^3$ can be deformed,
one to the other, by a deformation of $\R^3$.

If we assume that the isotopy $\phi: R^n\times I \rightarrow \R$
is differentiable, or smooth, then differentiating with
respect to $t$ produces a vector field on $\R^n$ and
the mapping $h = \phi(\cdot, 1)$ is obtained by integrating
this vector field from time $0$ to $1$, namely,
\[
\phi(x, 1) = x + \int_0^1 \frac{\partial d \phi(x, t)}{\partial t} \, dt ~,
\]
and $\partial d \phi(x, t)/\partial t$ is a vector
field on $\R^n$ for any given $t$.  However, this vector field
is time varying, since the partial derivative is not independent
of the time $t$.  This result shows that any two tame
embeddings of $S^2$ in $\R^3$ are connected by a flow along
a time-varying vector field from time $0$ to $1$.

\paragraph{Flow deformations. }
For our purposes, however, we desire the case where the two
embeddings are connected by a flow along a {\bf time-invariant}
vector field.  For a definition of flow, see the main paper,
and particularly the book \cite{lee2013smooth}, chapters 8 and 9.

We give here a sketch of a proof that this is possible under 
certain circumstances.

\begin{theorem}
If $S_0$ and $S_1$ are two smooth embeddings of $S^2$ in
$\R^3$ that do not intersect, then there is a vector field on
$\R^3$, which when integrated from $0$ to $1$ takes $S_0$
to $S_1$.  Otherwise stated, there is a flow on $\R^3$ that
takes $S_0$ at time $t=0$ to $S_1$ at time $t=1$.
\end{theorem}

\begin{proof}
Since $S_1$ and $S_0$ do not intersect, it follows that
$S_1$ must lie entirely inside or entirely outside of $S_0$.
Assume that it lies outside (otherwise reverse the roles of $S_0$
and $S_1$).

By the Generalized Schoenflies Theorem, 
there is a diffeomorphism $h$ of $R^3$ that takes $S_1$
to the standard sphere of radius $1$, which we shall denote by $S'_1$, and $S_0$ is mapped to
the interior of this sphere.  By a further diffeomorphism,
$S_0$ can be mapped to the sphere of radius $1/e$, which we denote
by $S_0'$, while leaving
the unit sphere fixed.  The composition of these two diffeomorphisms
is a diffeomorphism that takes $S_1$ to $S_1'$ 
and $S_0$ to $S_0'$.

Now, consider a flow on $R^3$ defined by 
$\Phi(x, t) = x e^t$.  The infinitessimal generator of this
flow%
\footnote{The {\em infinitessimal generator} of a flow is 
its derivative with respect to $t$, evaluated at $t=0$, namely
the vector field $V(x) = \Phi_x'(0)$. 
}
is given by $V(x) = x$, which is a smooth vector field.
This has the properties that
$\Phi_0(x) = x$ and $\Phi_1(x) = e x$.  Thus, the sphere of radius
$e$ is mapped to itself at time $0$ and to the sphere of radius $1$
at time $t=1$.

Now, we define the flow 
\begin{align}
\begin{split}
\label{eq:flow-transfer}
\Psi_t(x) &= h^{-1} \circ \Phi_t \circ h (x) \\
  &= h^{-1}(\Phi_t(h(x))) ~.
\end{split}
\end{align}
We verify that this is a flow on $R^3$ as follows.
Following from (\ref{eq:flow-transfer}) we have
\begin{align*}
\Psi_s \circ \Psi_t(x) 
  &= \Psi_s ( h^{-1}\circ \Phi_t \circ h(x)) \\
  &= h^{-1} \circ \Phi_s \circ h ( h^{-1} \circ \Phi_t\circ h(x)) \\
  &= h^{-1} \circ \Phi_s \circ \Phi_t \circ h(x) \\
  &= h^{-1} \circ \Phi_{s+t} \circ h(x) \\
  &= \Psi_{s+t} (x) ~
\end{align*}
This is the condition (see \cite{lee2013smooth}) that $\Psi(x, t)$ is a flow on $R^3$, and integration along the infinitessimal generator vector field
$V(x)$ defines the paths along which points flow.  Furthermore
at time $t=0$, the mapping $\Psi_t(x) = \Psi(x, t)$ is the identity map, taking $S_0$ to $S_0$,
and at time $t=1$, it maps $S_0$ to $S_1$.
The steps of this map are equal to

\[
S_0 \xrightarrow{h} S'_0 \xrightarrow{\Phi_1} S'_1
\xrightarrow{h^{-1}} S_1 ~.
\]

There is one extra case to be considered, that in which the
two spheres $S_0$ and $S_1$ are placed such that each one
is in the exterior part of the other sphere.  

The proof of this case is similar to the case previously considered.
In this case, one can show that there exists a diffeomorphism $h$ that
takes one sphere $S_0$ to the unit sphere centred at the origin $(0, 0, 0)$,
and the second sphere $S_1$ to the unit sphere $S_0'$ 
centred at point $(3, 0, 0)$.
Then a flow $\Phi(x, t) = x+3t$ is defined that takes $S_0'$ at time $0$
to $S_1'$ at time $t=1$.  Then  defining $\Psi$ as in 
\ref{eq:flow-transfer}, one verifies as before that $\Psi$ defines
a flow taking $S_0$ to $S_1$.
\end{proof}

Whether the theorem is true in general in the case where the two spheres
intersect, we are unable to determine, but our guess is that it
may not be true.

However, the restriction that the two embedded spheres $S_0$ and $S_1$
be disjoint is a harmless restriction for our purposes,
since a reconstruction of the surface can be placed at any point
in $\R^3$, in such a way that it does not intersect the standard
unit sphere.

\section{More Experiment Results}
\paragraph{Video demonstrator.} We have produced a short video clip containing re-rendered reconstructions under novel lighting, from novel viewpoints, and with novel materials (BRDFs). In the video, one can also visualize the training process, including both how the training loss reduces as well as how the intermediate reconstruction evolves as a function of epoch.  Notably, despite that both networks are trained from scratch, they offer effective supervision to each other through the minimizing of the loss function.  They converge quickly to a solution that is largely consistent with all input images. By contrast, traditional optimization-based methods often require a high quality initial shape (either from SFM or depth sensor) or an initial svBRDF.

\begin{figure*}[th!]
    \centering
    \includegraphics[width=1\textwidth]{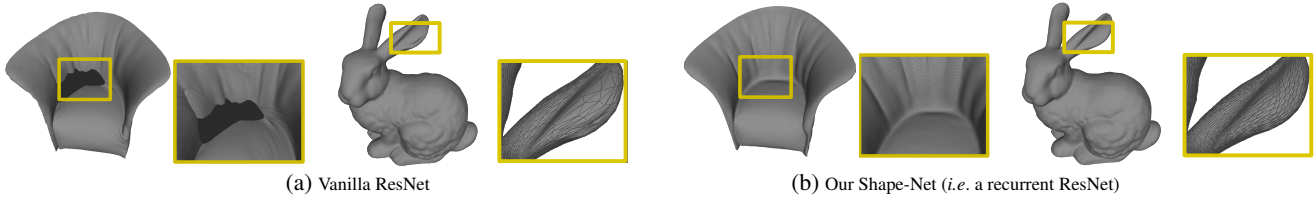}  
    \begin{subfigure}[b]{0.45\textwidth}\centering\caption{\scriptsize Vanilla ResNet}\end{subfigure}
    \begin{subfigure}[b]{0.45\textwidth}\centering\caption{\scriptsize Our Shape-Net (\ie a recurrent ResNet)}   \end{subfigure}  
    \caption{Comparing a vanilla ResNet (with no feedback loop) with our Shape-Net model (\ie a ResNet with a feedback ring connection). The vanilla ResNet leads to many back-face and self-intersections, while our ShapeNet is free from self-intersection, and yields visually more pleasant watertight reconstruction. {\bf Better viewed on screen with zoom in.}}
    \label{fig:shapenet_vs_baseline}
\end{figure*}

\paragraph{Ablation study of a plain (non-recursive) ResNet.} Neural networks in general, however, do not warrant these properties. Fig-\ref{fig:shapenet_vs_baseline} illustrates the object surface recovered by a non-recursive ResNet structure versus our Shape-Net model. We observe that our Shape-Net structure creates an intersection-free surface, while baseline model suffers self-intersection and flipping-over (dark regions) issues. 

\paragraph{Refinement of Camera Pose/Light Position.} 
In all our previous experiments so far, we have assumed the camera pose and light source position are pre-calibrated. In fact, within the same framework of our neural solver, we are able to refine camera and light as well (at least to refine their initial calibrations). To validate this idea, we back-propagate error to the 6-DOF camera pose, and obtain the results in Tab-\ref{tab:camera_pose}, which confirm that calibration errors are reduced by almost half. Similar improvements were obtained for refining light-source position.
\begin{table}
    \centering
    \begin{tabular}{c*3{|c}}
        Pose &  Normal (degs) & Depth \% & PSNR (dB)\\\hline
        Refined & 6.60 & 0.65 & 32.9\\\hline
        Baseline & 10.3 & 1.34 & 25.6 \\\hline
    \end{tabular}
    \caption{\small By back-propagating training loss to camera pose, we can also refine camera calibration accuracy. Here shows the mean normal and depth errors and image PSNR error on \textit{Bunny} starting from a poor initial calibration.}
    \label{tab:camera_pose}
\end{table}

\subsection{Multiple level of details} 
Because our Shape-Net implements a continuous diffeomorphic mapping, in principle it allows one to reconstruct a shape at any arbitrary resolution (up to the level of details that the Neural Nets have learned during training).

\begin{figure*}[!h!]  
    \centering
    \includegraphics[width=\textwidth, trim={0cm 0cm 0cm 0cm}, clip]{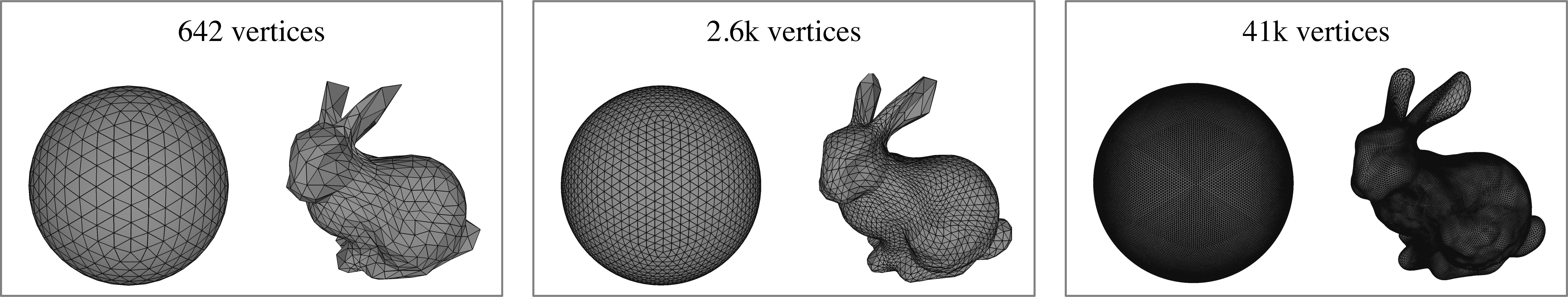}
    \caption{Our method produces a continuous surface reconstruction, which allows one to obtain a watertight mesh reconstruction at any resolution (up to the level of details learned by the neural networks). {\bf Better viewed on screen.}} 
    \label{fig:sample_mesh}
\end{figure*}

An example is shown in Fig-\ref{fig:sample_mesh}, where we change the sampling resolution of meshed surface (\textit{Bunny}) from Shape-Net by changing the number of vertices.
  Compared with implicit surface representation (\cf DeepSDF~\cite{park2019deepsdf}), a meshed surface can be easily rendered by rasterization, hence is more efficient than ray-tracing.

\begin{figure*}[!h] 
    \centering
    \includegraphics[width=\linewidth]{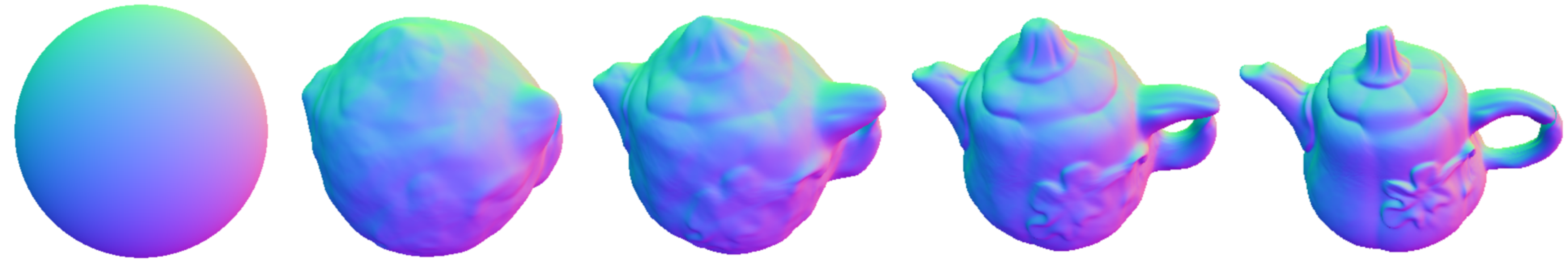}
    \caption{Illustration of how a genus-1 teapot is approximated by a genus-0 embedding with our method. Note the handle of the teapot is elegantly broken, making it a valid genus-0 shape.}
    \label{fig:deform_pot}
\end{figure*}
\subsection{Approximating higher-genus by genus-zero} We show how higher-genus shapes can be approximated by a genus-0 sphere embedding for the teapot model. Interestingly, our method is able to reconstruct the target shape despite of a incorrect surface topology.  Fig-\ref{fig:deform_pot} illustrates how the shape evolves (as a function of the time $t$) from the initial sphere to the target object.

\begin{figure}[!h]
    \centering
    \includegraphics[width=0.5\textwidth]{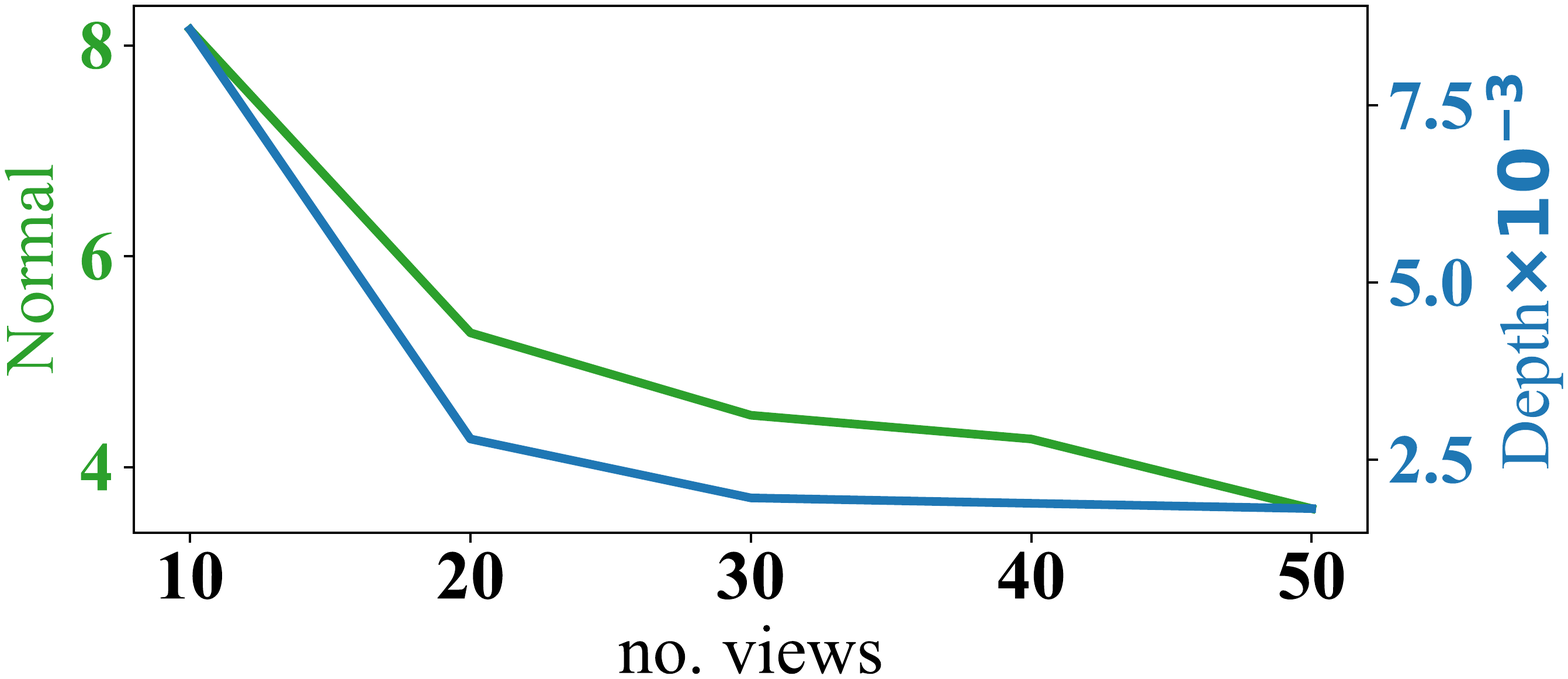}
    \caption{Errors versus Number of input views tested on \textit{Bunny}. The blue curve shows errors of depth estimation, and the green curve shows errors in surface normal estimation.}
    \label{fig:no_views}
\end{figure}

\subsection{Performance versus Number of views}  We also provide comparisons of performances versus different numbers of input views.  Specifically, we evaluate the accuracy of shape as the number of views changes. The results are shown in Fig-\ref{fig:no_views}. We observe that while a larger number of views better constrain shape and reflectance, the accuracy does not drop significantly \wrt number of views until less than 30 views are given. 
\paragraph{Recovering high-frequency details.} Despite the use of positional encoding, we noted that for certain challenging shape our method has missed some small geometric details. We believe this is due to two reasons: 1. Our small-sized network has limited expressive power.  In fact, empirically, we notice our Shape-Net parameterization only takes up 1/5th of the RAM space of the mesh parameterization for the same surface used in the image rendering stage. This suggests that the expressiveness of our new method can be further improved if we increase the size of the network.  2. Input images are of limited resolution.  We currently only used $512\times512$ images, due to the memory limitation of our single GPU.  

We believe increasing network size and image resolution will be able to further improve the quality of recovered shape. Even with a small network, our method achieved much better quantitative reconstruction, and the overall visual quality is arguably much better too, compared with other competing methods.
\subsection{Timing Comparison} We focus this paper on the framework and algorithm aspects, rather than computational efficiency. At least it was not our top priority in developing this paper. However, we note our method is reasonably fast, compared with both traditional photometric methods and new deep learning based shape reconstruction methods. For example, our method took about 4--8 hours in training on one scene based on a single GPU, and its testing time (and re-rendering time) is almost instant.  In contrast, the initialization alone in Park's paper already requires several hours, and NeRF was reported to spend about 10--20 hours to train a complex scene.  Furthermore, many traditional photometric methods relying on special hardware which consume hours just for image acquisition, while our method only requires a smartphone camera with an inbuilt flashlight (used as the active light source) for image acquisition.  

\subsection{More Relighting and View Synthesis results}
\begin{figure*}[!h]
    \centering
    \includegraphics[width=\textwidth]{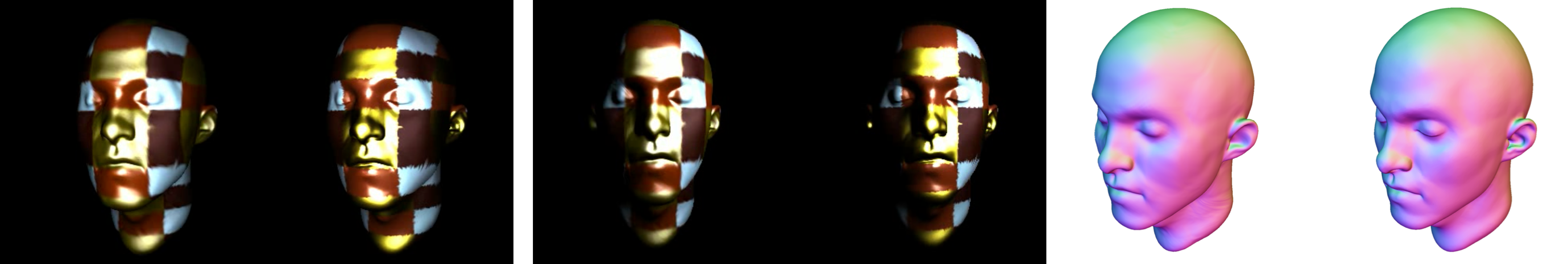}
    
    \begin{subfigure}[b]{0.15\textwidth}
    \centering\caption*{Ours}
    \end{subfigure}
    \begin{subfigure}[b]{0.15\textwidth}
    \centering\caption*{G.T.}
    \end{subfigure}
    \begin{subfigure}[b]{0.15\textwidth}
    \centering\caption*{Ours}
    \end{subfigure}
    \begin{subfigure}[b]{0.15\textwidth}
    \centering\caption*{G.T.}
    \end{subfigure}
    \begin{subfigure}[b]{0.15\textwidth}
    \centering\caption*{Ours}
    \end{subfigure}
    \begin{subfigure}[b]{0.15\textwidth}
    \centering\caption*{G.T.}
    \end{subfigure}
    \begin{subfigure}[b]{0.30\textwidth}
    \centering\caption{ Novel view synthesis}
    \end{subfigure}
    \begin{subfigure}[b]{0.30\textwidth}
    \centering\caption{ Relighting}
    \end{subfigure}
    \begin{subfigure}[b]{0.30\textwidth}
    \centering\caption{ Normal and shape}
    \end{subfigure}
    \caption{Re-rendered images and shape of \textit{Head} model under novel viewpoints and novel light (left), compared with the ground truth (right). Please see the demo video for better visualization.}
    \label{fig:head}
\end{figure*}

\begin{figure*}
\centering
\includegraphics[width=0.9\textwidth]{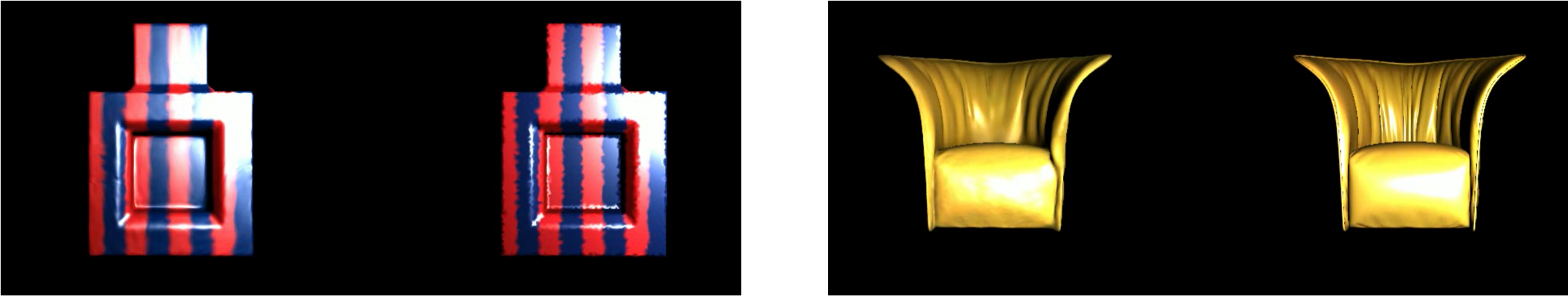}
\begin{subfigure}[b]{0.22\textwidth}
\centering\caption*{Ours}
\end{subfigure}
\begin{subfigure}[b]{0.22\textwidth}
\centering\caption*{G.T.}
\end{subfigure}
\begin{subfigure}[b]{0.04\textwidth}
\centering\caption*{}
\end{subfigure}
\begin{subfigure}[b]{0.22\textwidth}
\centering\caption*{Ours}
\end{subfigure}
\begin{subfigure}[b]{0.22\textwidth}
\centering\caption*{G.T.}
\end{subfigure}
\includegraphics[width=0.9\textwidth]{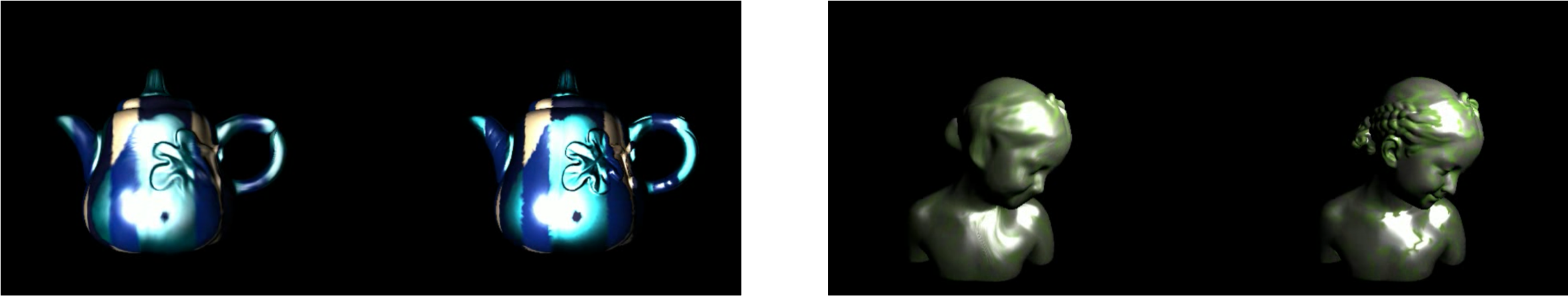}
\begin{subfigure}[b]{0.22\textwidth}
\centering\caption*{Ours}
\end{subfigure}
\begin{subfigure}[b]{0.22\textwidth}
\centering\caption*{G.T.}
\end{subfigure}
\begin{subfigure}[b]{0.04\textwidth}
\centering\caption*{}
\end{subfigure}
\begin{subfigure}[b]{0.22\textwidth}
\centering\caption*{Ours}
\end{subfigure}
\begin{subfigure}[b]{0.22\textwidth}
\centering\caption*{G.T.}
\end{subfigure}
\caption{Re-rendered images under novel lights.}
\label{fig:novel_light}
\end{figure*}

\begin{figure*}
\centering
\includegraphics[width=0.9\textwidth]{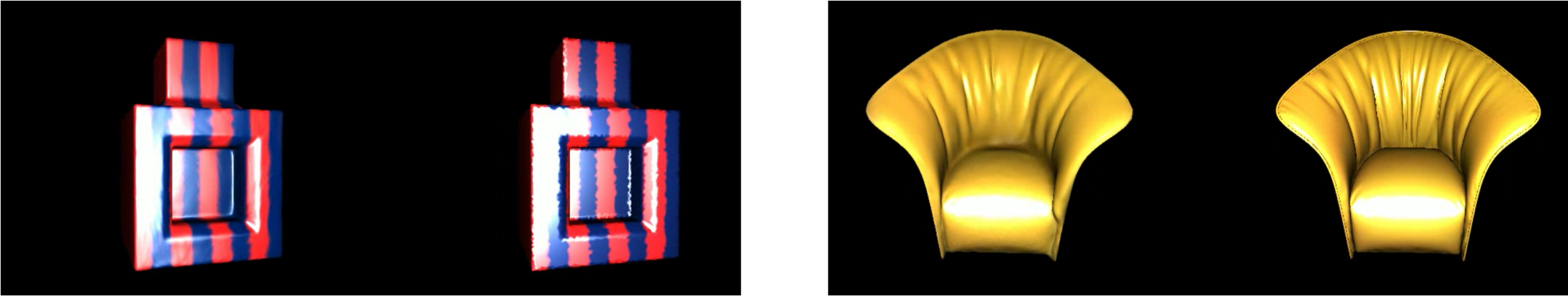}
\begin{subfigure}[b]{0.22\textwidth}
\centering\caption*{Ours}
\end{subfigure}
\begin{subfigure}[b]{0.22\textwidth}
\centering\caption*{G.T.}
\end{subfigure}
\begin{subfigure}[b]{0.04\textwidth}
\centering\caption*{}
\end{subfigure}
\begin{subfigure}[b]{0.22\textwidth}
\centering\caption*{Ours}
\end{subfigure}
\begin{subfigure}[b]{0.22\textwidth}
\centering\caption*{G.T.}
\end{subfigure}
\includegraphics[width=0.9\textwidth]{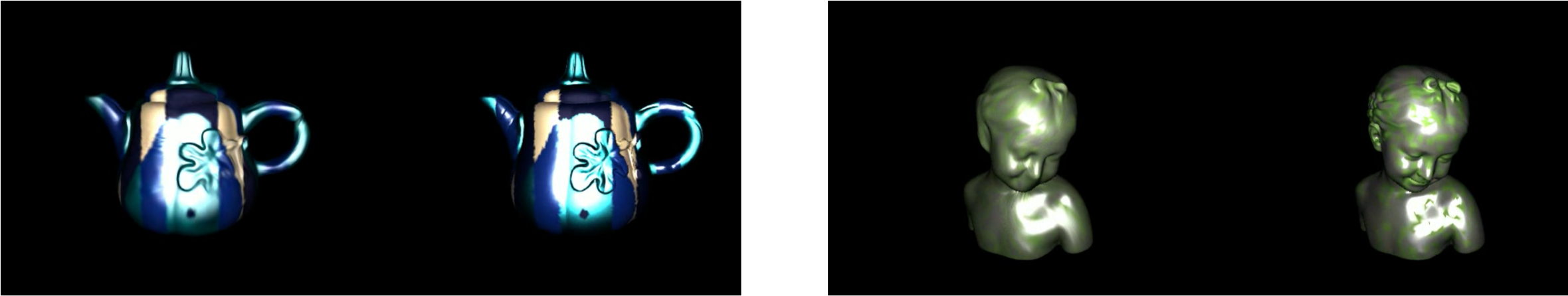}
\begin{subfigure}[b]{0.22\textwidth}
\centering\caption*{Ours}
\end{subfigure}
\begin{subfigure}[b]{0.22\textwidth}
\centering\caption*{G.T.}
\end{subfigure}
\begin{subfigure}[b]{0.04\textwidth}
\centering\caption*{}
\end{subfigure}
\begin{subfigure}[b]{0.22\textwidth}
\centering\caption*{Ours}
\end{subfigure}
\begin{subfigure}[b]{0.22\textwidth}
\centering\caption*{G.T.}
\end{subfigure}
\caption{Re-rendered images from Novel viewpoints.}\label{fig:novel_view}
\end{figure*}

Because our method is essentially based on inverse rendering (inverse graphics), it naturally allow to generate re-rendered images under different lights, and from different view-points (\ie novel view synthesis). ~Fig-\ref{fig:head} shows the recovered object under different lights, and from different viewpoints, compared with the corresponding ground truth images.  Fig-\ref{fig:novel_light} demonstrates the recovered object under different, non-co-located lighting conditions versus the corresponding ground truth images.   ~Fig-\ref{fig:novel_view} illustrates the recovered object rendered under novel viewpoints,  compared with the corresponding ground truth images.

\section{Reproducible Research} 
Our method is simple, requiring only two small MLPs as the backbone networks.  Due to this {\em remarkably simple} design, our experiment results is easy to reproduce. In fact, the core of our method is implemented in less than 200 lines of Python code. Note also the Soft Rasterizer is already supported by the PyTorch-3D Library. We will release all our source codes and models for facilitating reproducible research.

\section{Broader Impact.} We expect this paper will have broader impact to the research community. It solves an open challenging problem of multi-view reconstruction of  complex 3D geometry and unknown materials of the physical world from  easily accessible 2D pictures of it.  In particular, our method achieves high quality 3D modelling of objects with unknown arbitrary, and spatially-varying non-Lambertian surface reflectances using only standard multi-view observations.  Applications of the method could be anywhere as long as 3D information is required.  This could be the case in: product design, e-commerce, virtual and augmented reality, entertainment, security, medical imaging, autonomous driving and more.  


